\UseRawInputEncoding
\pdfoutput=1

\documentclass[11pt]{article}

\usepackage{acl}
\usepackage{fix-cm}
\usepackage{times}

\usepackage{microtype}
\usepackage{float} % For [H] float placement
\usepackage{graphicx}   % For including images
\usepackage{caption}    % For customizing captions
\usepackage{multirow}
\usepackage{tabularx}
\usepackage{booktabs}
\definecolor{citecolor}{HTML}{0071BC}
\definecolor{linkcolor}{HTML}{ED1C24}
\definecolor{LGray}{gray}{0.97}
\usepackage{multicol}
\usepackage{colortbl}
\usepackage{xcolor}
\definecolor{tabhighlight}{HTML}{e5e5e5}
\usepackage{color}
\usepackage{xspace}
\usepackage{pifont}
\usepackage{colortbl} 
\usepackage{subcaption}
\usepackage{placeins} 
\usepackage{tcolorbox}
\tcbuselibrary{listings}
\usepackage{listings}
\usepackage{cuted} 

\definecolor{Teal}{HTML}{4B959C}     
\definecolor{pinky}{HTML}{F092D2}    
\definecolor{SkyBlue}{HTML}{62C1CA}     
\definecolor{yel}{HTML}{F3BB23}     
\definecolor{ppl}{HTML}{93358F}
\definecolor{grn}{HTML}{64C9AE}          
\definecolor{highlightcyan}{RGB}{150, 255, 255}

\usepackage{graphicx}
\usepackage{tikz}
\usetikzlibrary{trees}

\definecolor{darkgreen}{rgb}{0.0, 0.5, 0.0}  % 
\newcommand{\dmark}{\textcolor{gray!40}{\textbf{--}}} 

\definecolor{olivegreen}{rgb}{0.2,0.5,0.3}
\definecolor{custombrown}{rgb}{0.7,0.1,0.1}

\newcommand{\cmark}{\textcolor{olivegreen}{\scalebox{1}[1.0]{\ding{51}}}} 
\newcommand{\xmark}{\textcolor{custombrown}{\ding{55}}}                          % Red X

\lstdefinelanguage{json}{
    basicstyle=\ttfamily\scriptsize,
    numbers=left,
    numberstyle=\tiny,
    stepnumber=1,
    breaklines=true,
    showstringspaces=false,
    frame=single,
    backgroundcolor=\color{gray!10},
    keywordstyle=\color{blue},
    stringstyle=\color{red}
}

\usepackage[T1]{fontenc}
\usepackage[utf8]{inputenc}
\usepackage{inconsolata}
\usepackage{graphicx}

\title{
    \begin{minipage}{0.12\textwidth} 
        \raggedleft
        \includegraphics[height=1.75cm]{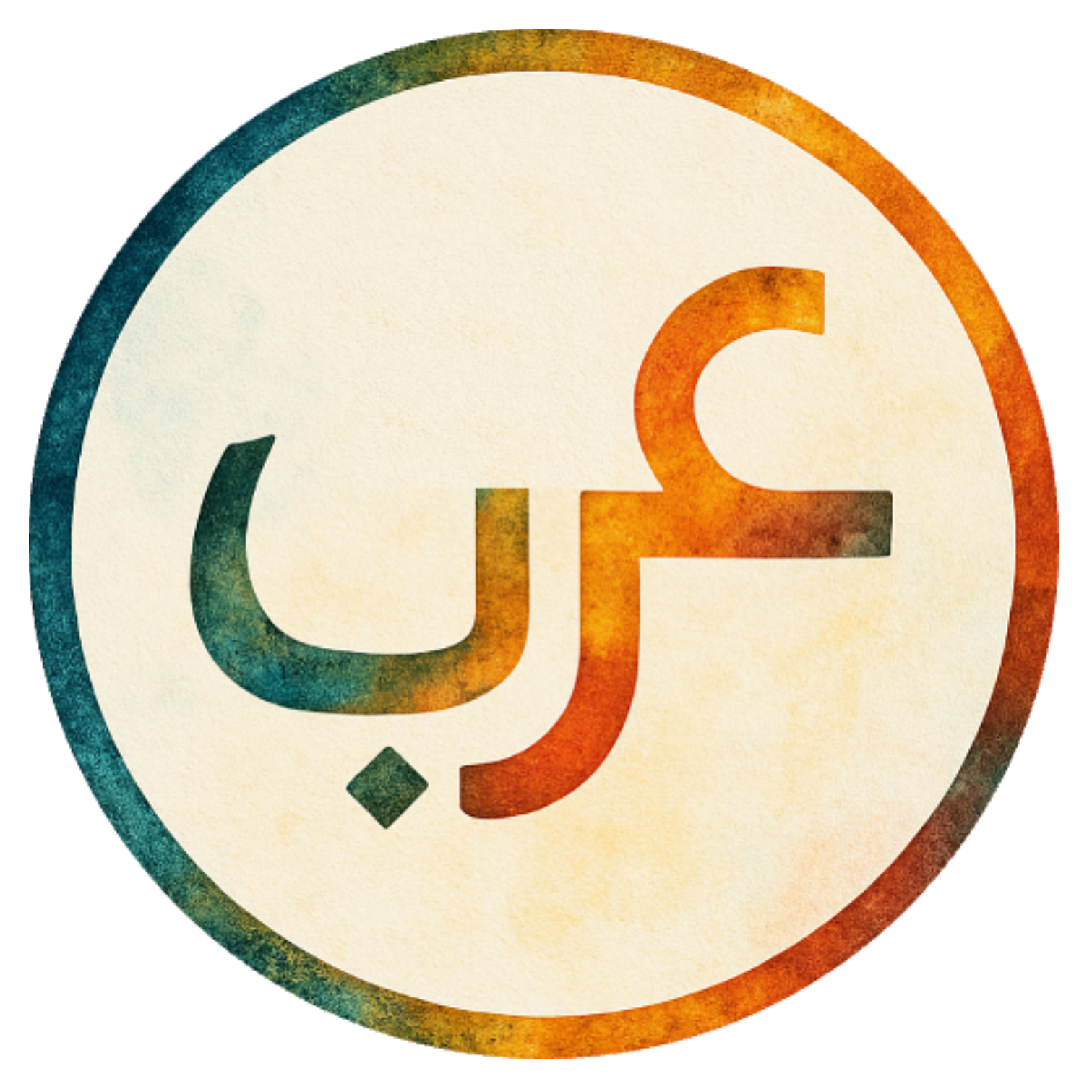} 
    \end{minipage}%
    \hspace{0.0002cm} 
    \begin{minipage}{0.8\textwidth} 
        \centering
        \textbf{ARB: A Comprehensive Arabic Multimodal Reasoning Benchmark}
    \end{minipage}
}

\author{\\  \quad{Sara Ghaboura}\textsuperscript{1$\dagger$} \quad {Ketan More}\textsuperscript{1$\dagger$} \quad {Wafa Alghallabi}\textsuperscript{{1}}  \quad{Omkar Thawakar}\textsuperscript{1} \\
\quad{Jorma Laaksonen}\textsuperscript{3} \quad 
    {Hisham Cholakkal}\textsuperscript{1} \quad  
     {Salman Khan}\textsuperscript{1,2} \quad  {Rao Muhammad Anwer}\textsuperscript{1,3}\\
     \fontsize{11pt}{12pt}\selectfont \textsuperscript{1}Mohamed bin Zayed University of AI,  \textsuperscript{2}Australian National University,
     \textsuperscript{3}Aalto University \\
     \fontsize{10pt}{12pt}\selectfont \{{sara.ghaboura, ketan.more}\}@mbzuai.ac.ae \\
 {\hypersetup{urlcolor=blue}
\fontsize{11pt}{12pt}\selectfont \href{https://mbzuai-oryx.github.io/ARB/}{https://mbzuai-oryx.github.io/ARB/}}}

\begin{document}
\maketitle

\begin{abstract}
As Large Multimodal Models (LMMs) become more capable, there is growing interest in evaluating their reasoning processes alongside their final outputs. However, most benchmarks remain focused on English, overlooking languages with rich linguistic and cultural contexts, such as Arabic. To address this gap, we introduce the Comprehensive Arabic Multimodal Reasoning Benchmark (ARB), the first benchmark designed to evaluate step-by-step reasoning in Arabic across both textual and visual modalities. ARB spans 11 diverse domains, including visual reasoning, document understanding, OCR, scientific analysis, and cultural interpretation. It comprises 1,356 multimodal samples paired with 5,119 human-curated reasoning steps and corresponding actions. We evaluated 12 state-of-the-art open- and closed-source LMMs and found persistent challenges in coherence, faithfulness, and cultural grounding. ARB offers a structured framework for diagnosing multimodal reasoning in underrepresented languages and marks a critical step toward inclusive, transparent, and culturally aware AI systems. We release the benchmark\footnote{\href{https://huggingface.co/datasets/MBZUAI/ARB}{https://huggingface.co/datasets/MBZUAI/ARB}}, rubric, and evaluation suit \footnote{\href{https://github.com/mbzuai-oryx/ARB}{https://github.com/mbzuai-oryx/ARB}} to support future research and reproducibility.
\def\thefootnote{$\dagger$}\footnotetext{Equal contribution.}
\end{abstract}

\begin{figure*}[t!]
\centering  
\includegraphics[width=\textwidth,height=10cm]{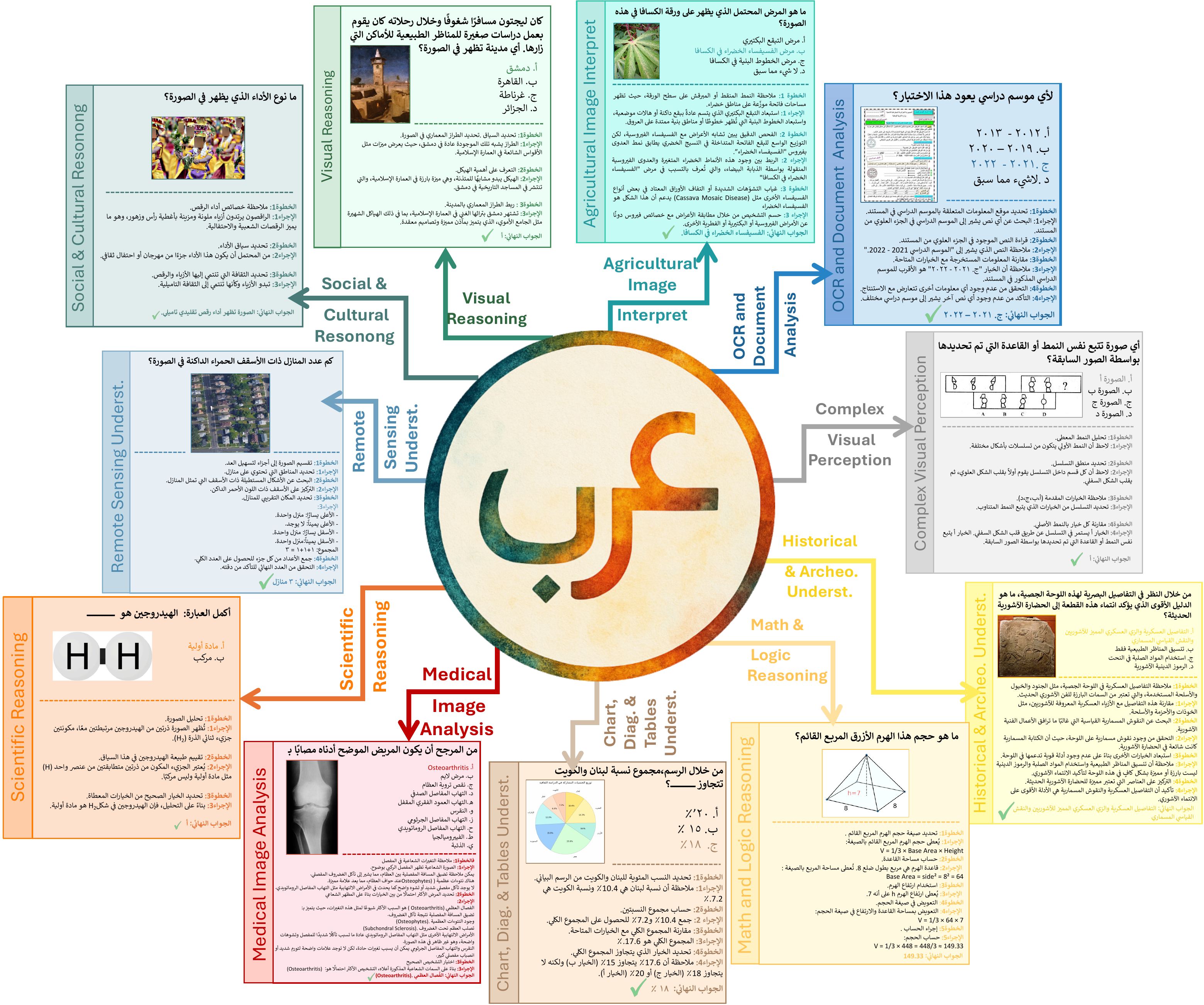}
\vspace{-1.5em}
\caption{\textbf{ARB Dataset Diversity.} ARB comprises a wide array of multimodal reasoning samples, each combining a visual input with an Arabic question and detailed step-by-step reasoning with actions taken by step. The dataset spans 11 distinct domains, including visual reasoning, OCR and document understanding, chart and diagram interpretation, mathematical and logical inference, scientific and medical analysis, cultural and historical interpretation, remote sensing, agricultural image analysis, and complex visual perception—capturing the linguistic richness, cultural depth, and cross-domain complexity essential for evaluating reasoning in Arabic.}
\label{fig:samples_intro}
\vspace{-0.5em}
\end{figure*}   

\section{Introduction}
Arabic, spoken by more than 400 million people worldwide, embodies significant linguistic diversity and a profound cultural heritage. Despite its widespread usage, Arabic remains notably underrepresented in advanced AI systems, particularly those that involve multimodal reasoning, simultaneous interpretation, and logical processing of textual and visual data crucial for fields such as education, healthcare, and cultural preservation. This scarcity limits the deployment and inclusion of multimodal AI in Arabic-speaking communities.

\begin{figure*}[hptb]
\centering  
\includegraphics[width=\textwidth, height=3.75cm]{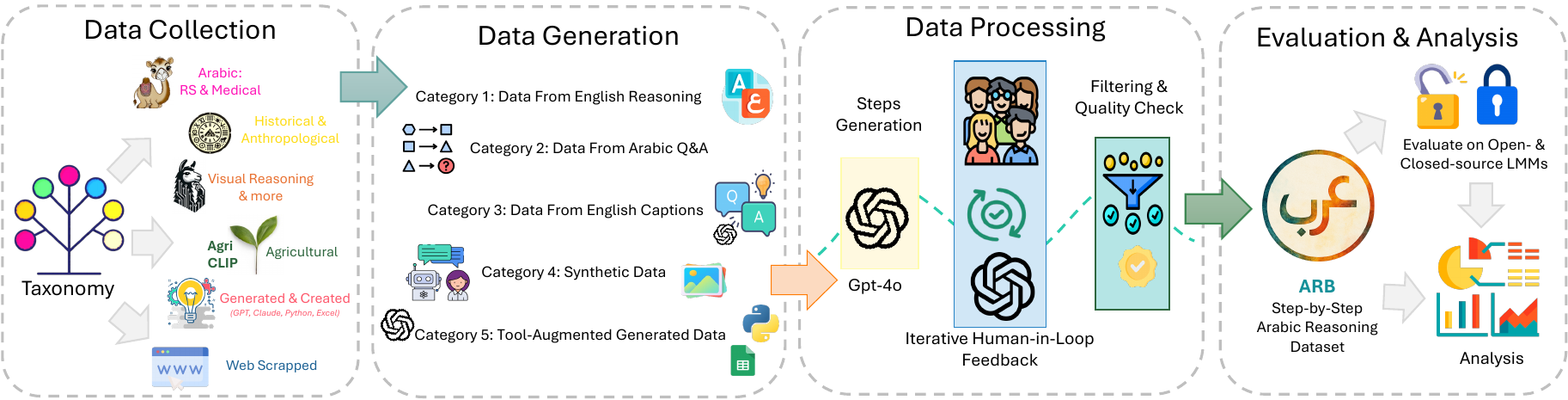}
\vspace{-1.5em}
\small
\caption{\textbf{The ARB Dataset Pipeline.} The figure illustrates the ARB pipeline for evaluating Arabic multimodal reasoning in LMMs. It begins with data collection across 11 domains—such as medical imaging, historical interpretation, visual reasoning, and agriculture—sourced from curated datasets (e.g., VRC-Bench, CAMEL-Bench), synthetic content, tool-augmented outputs, and web scraping. Data is generated across five categories: English reasoning chains, Arabic Q\&A, English captions, synthetic samples, and tool-enhanced content. Reasoning steps are refined via human-in-the-loop feedback and filtered for logical consistency and cultural alignment. The benchmark supports fine-grained evaluation of open- and closed-source models on Arabic step-by-step reasoning.}

\label{fig:camel_o1_pipeline}
\vspace{-1em}
\end{figure*}

Recent developments in LMMs reflect a growing emphasis on transparency and interpretability, achieved through explicit reasoning steps. Techniques such as chain-of-thought (CoT) prompting, initially introduced by \citet{wei2022chain}, encourage models to systematically articulate intermediate reasoning steps, significantly improving both performance and explainability. This paradigm has gained traction in English-based language models and has been effectively extended to multimodal settings in models such as LLaVA-CoT \cite{xu2025llavacotletvisionlanguage}, VisCoT \cite{shao2024visual}, and the recent LLamaV-o1 \cite{thawakar2025llamav}.

Current step-by-step reasoning benchmarks largely focus on English, overlooking the linguistic nuances and cultural contexts essential to Arabic. Recent work on cross-lingual reasoning \cite{yong2025crosslingual} shows that English-trained models can generalize to other languages via test-time scaling; however, Arabic was not explicitly evaluated, and performance often falters in the presence of linguistic complexity and cultural commonsense. Existing Arabic multimodal data sets, such as CAMEL-Bench \cite{ghaboura2024camel}, Henna \cite{alwajih2024peacock}, and JEEM \cite{kadaoui2025jeem}, prioritize final answer accuracy with limited attention to intermediate reasoning. Meanwhile, benchmarks like AraDiCE \cite{mousi2024aradice} and ArabCulture \cite{sadallah2025commonsense} remain confined to textual modalities. Together, these limitations signal the need for Arabic-specific multimodal reasoning benchmarks that reflect the linguistic and cultural demands of the target language.

To bridge this critical chasm, we introduce the Comprehensive Arabic Multimodal Reasoning Benchmark (ARB), the first explicitly designed benchmark for evaluating detailed step-by-step reasoning in Arabic multimodal contexts (Table~\ref{tab:ours_vs_others}). ARB comprises 1,356 multimodal samples in 11 domains, including visual reasoning, document understanding, optical character recognition (OCR), cultural interpretation, medical imaging, and remote sensing (Figure~\ref{fig:samples_intro}). Each sample includes meticulously curated annotations with more than 5.1k reasoning steps, each paired with a specific action, allowing nuanced assessment of coherence, fidelity, and cultural grounding beyond mere final-answer accuracy.

\begin{table}[t!]
    \centering
    \setlength{\tabcolsep}{7pt}
    \renewcommand{\arraystretch}{1.15}
    \resizebox{0.49\textwidth}{!}{%
    \begin{tabular}{lcccccc}
    \toprule
    \cellcolor{gray!10}& \cellcolor{gray!10}\textbf{Multi-} &\cellcolor{gray!10}\textbf{Multi-}& \cellcolor{gray!10}\textbf{Reasoning}&\cellcolor{gray!10}\textbf{Open-}&\cellcolor{gray!10}\textbf{Eval.}\\ 
     
   \rowcolor{gray!10} \multirow{-2}{*}{\textbf{Benchmarks}}&\textbf{modal?}&\textbf{domain?}   & \textbf{support?} &\textbf{source?}&\textbf{Level} \\  
    \midrule
    Henna        &\cmark  & \xmark   & \xmark & \xmark&  \textcolor{custombrown}{FA\textsuperscript{*}}\\
    CAMEL-Bench  & \cmark & \cmark   & \xmark & \cmark & \textcolor{custombrown}{FA\textsuperscript{*}} \\
    AraSTEM      & \xmark & \xmark   &\cmark & \cmark &  \textcolor{custombrown}{FA\textsuperscript{*}}\\
    AraDiCE      & \xmark & \cmark   &\cmark  & \cmark  &  \textcolor{custombrown}{FA\textsuperscript{*}}\\
    JEEM         & \cmark & \cmark   &\xmark  &  \cmark & \textcolor{custombrown}{FA\textsuperscript{*}}\\ 
    PALM         & \xmark & \cmark   &\xmark & \xmark &  \textcolor{custombrown}{FA\textsuperscript{*}}\\ 
    ArabCulture  & \xmark & \cmark   &\cmark  & \cmark & \textcolor{custombrown}{FA\textsuperscript{*}}\\ 
   \toprule
  \textbf{ARB (ours)} & \cmark &  \cmark & \cmark& \cmark& \textbf{\textcolor{olivegreen}{FA\textsuperscript{*}\& Step\textsuperscript{*}}} \\
   \bottomrule
    \end{tabular}}
    \vspace{-0.5em}
   \caption{\textbf{Comparison of our ARB with existing Arabic LMM benchmarks and Reasoning Benchmarks.} \small{ Henna \cite{alwajih2024peacock}, CAMEL-Bench \cite{ghaboura2024camel}, AraSTEM \cite{mustapha2024arastem}, AraDiCE \cite{mousi2024aradice}, JEEM \cite{kadaoui2025jeem}, PALM \cite{alwajih2025palm}, ArabCulture \cite{sadallah2025commonsense}}. FA\textsuperscript{*}: Final Answer Evaluation. Step\textsuperscript{*}: Step-level Evaluation. }
   \vspace{-1.5em}
    \label{tab:ours_vs_others}
    \end{table}

The construction of ARB involved systematic identification of critical reasoning domains and rigorous data sourcing, validated by domain experts. All annotations, reasoning chains, and actions were verified by native speakers through a human-in-the-loop process to ensure logical precision and cultural fidelity. We also performed a human evaluation to assess the correctness of the reasoning steps and to validate the reliability of using LLMs as automated judges.

Evaluations of 12 prominent open-source and closed-source LMMs - including GPT-4V \cite{openai2024gpt4ocard,openai2024gpt4omini,openai2025gpt41,openai2025o3o4}, Gemini variants \cite{google2024gemini15pro,google2024gemini20flashthinking}, and open-source multilingual models such as Qwen2.5-VL \cite{qwen2025vl}, LlaMA variants \cite{meta2024llama32vision,meta2025llama4scout}, Aya-Vision \cite{cohere2025aya8b}, InternVL3 \cite{chen2024expanding}, and Arabic-focused AIN \cite{heakl2025ain} - highlight significant deficiencies in Arabic reasoning coherence and cultural grounding despite robust English performance, underscoring the necessity of ARB.

In summary, (1) we introduce ARB, the first Arabic-centric benchmark designed to evaluate step-by-step multimodal reasoning across 11 culturally and linguistically grounded domains; (2) we conduct extensive evaluations of 12 leading open- and closed-source LMMs, uncovering limitations in coherence, faithfulness, and reasoning quality in Arabic; (3) we integrate a human-in-the-loop pipeline with manual verification by native speakers and domain experts to ensure annotation accuracy; and (4) we perform human evaluations to validate reasoning correctness and assess the effectiveness of LLM-as-a-judge scoring.

\vspace{-0.5em}

\begin{figure*}[t!]
\centering  
\includegraphics[width=\textwidth,height=7.25cm]{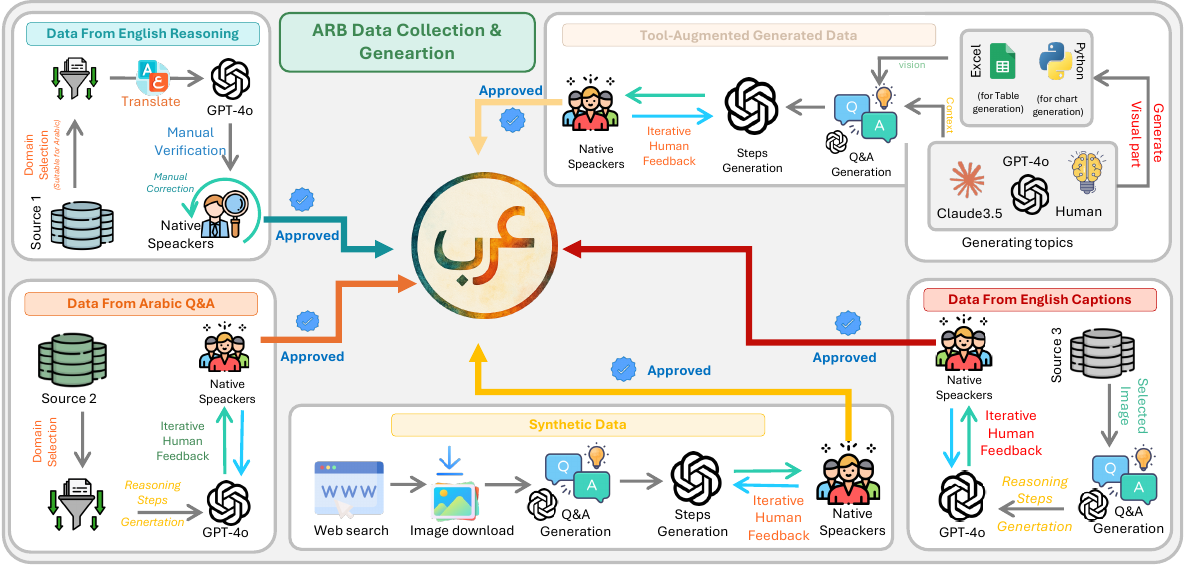}
\vspace{-1.5em}
  \caption{\textbf{Overview of the ARB Data Collection, Generation and Verification Framework.} The ARB benchmark is constructed from five primary data sources: (1) English reasoning benchmarks, (2) Arabic question–answer benchmarks, (3) English-captioned datasets, (4) Synthetic data, and (5) Tool-augmented data. All data undergoes iterative refinement through human-in-the-loop feedback and validation by native Arabic speakers to ensure cultural and linguistic fidelity.}
  \label{fig:data_collection_gen_verify}
  \vspace{-0.75em}
\end{figure*}

\section{Step-by-Step Arabic Reasoning Benchmark: ARB}
Figure~\ref{fig:camel_o1_pipeline} presents an overview of the ARB data construction pipeline, which we describe in detail through the following subsections.
\vspace{-.5em}
\subsection{Data Collection}
We adopt a domain-guided approach to curate data across a broad spectrum of categories relevant to Arabic multimodal reasoning. This ensures diversity in both content and modality, encompassing textual and visual tasks. The selected domains (Figure~\ref{fig:samples_intro})—from visual perception to historical and anthropological interpretation—are sourced from existing benchmarks, human-authored questions, and synthetic content (Table~\ref{tab:data_sources}). These sources were selected to capture diverse reasoning challenges and promote linguistic, cognitive, and cultural variety across the dataset.

\begin{table}[t!]
    \centering
    \setlength{\tabcolsep}{3pt}
    \resizebox{0.47\textwidth}{!}{%
    \begin{tabular}{lcccc}
    \hline
    \multirow{2}{*}{\textbf{Domains}} & \textbf{English} & \textbf{Arabic} & \textbf{Human} & \multirow{2}{*}{\textbf{Synthetic}}\\  
    & \textbf{Bench} & \textbf{Bench} & \textbf{Created} & \\  
    \hline
    Visual Reasoning & \cmark &\dmark &\dmark & \dmark\\
    OCR \& Docs Anal. &\dmark  &\dmark& \cmark & \cmark\\
    CDT & \cmark & \cmark & \cmark & \cmark \\
    Math \&logic & \cmark &\dmark &\dmark &\dmark \\
    Social \& Cult.& \cmark & \dmark &\dmark &\dmark\\
    Comp. Vis. Percept. & \cmark & \dmark&\dmark& \dmark \\
    Medica Img. Anal. & \cmark & \cmark &\dmark &\dmark\\
    Scientific Reasoning & \cmark &\dmark& \dmark &\dmark  \\
    Agricultural Interp. & \cmark &  \dmark& \cmark & \cmark\\
    Remote Sensing Und.&  \dmark& \cmark &\dmark &\dmark \\
    Histo. \& Anthro. & \cmark &\dmark & \cmark & \cmark \\
   \bottomrule
    \end{tabular}}
    \vspace{-0.5em}
   \caption{\textbf{Source Types Across ARB Domains.} We show the sources for each of the 11 domains, indicating whether data originated from Arabic or English benchmarks, human-written questions, or synthetic content, highlighting the dataset’s linguistic and cognitive diversity. \tiny {\textbf{CDT:} Chart, Diagrams, \& Table Understanding; \textbf{Social \& Cult.:} Social \& Cultural Reasoning; \textbf{Complex Vis. Percept.}: Complex Visual Perception;  \textbf{Agricultural Interp.:} Agricultural Image Interpretation; \textbf{Histo. \& Anthro.}: Historical \& Anthropological Understanding.}}
   \vspace{-1.25em}
    \label{tab:data_sources}
    \end{table}

\subsection{Data Generation and Data Processing}
We generated the dataset content in five main categories, each targeting a different source or creation method (Figure~\ref{fig:data_collection_gen_verify}). For each category, we employed a strategically selected prompting technique and engaged human experts to iteratively review and refine the resulting reasoning steps.

\noindent\textbf{Category 1: English Reasoning Benchmarks}\\
We adapted the English step-by-step reasoning dataset VRC-Bench \cite{thawakar2025llamav} by excluding domains with limited Arabic relevance (e.g., OCR, Charts, Diagrams \& Tables). The remaining content was translated into Arabic using GPT-4o and reviewed by native speakers for step-level accuracy, coherence, and fluency. Particular attention was given to resolving translation challenges such as singular–plural and subject–verb agreement, sentence structure differences, and non-literal expressions. Figurative language and cultural references were carefully localized to preserve contextual relevance, meaning complexity, and naturalness in Arabic.

\noindent\textbf{Category 2: Arabic QA Benchmarks }\\
To further enrich the ARB collection, we incorporate two specialized domains, medical image analysis and remote sensing understanding, sourced from the CAMEL-Bench \cite{ghaboura2024camel}. For each QA pair, we generated detailed step-by-step reasoning traces to support interpretability and structured inference using GPT-4o. For the medical domain, we employed a \textbf{few-shot CoT prompting} strategy to produce coherent reasoning chains. However, this approach proved insufficient for the remote sensing domain, where questions often require spatial decomposition and complex visual inference. To address this, we adopted the \textbf{plan-and-solve prompting} framework \cite{wang2023plan}, guiding the model to divide images into segments (e.g., quadrants or longitudinal zones) and apply a structured, divide-and-conquer reasoning approach. This significantly improved the fidelity and completeness of reasoning in the remote sensing domain.

\noindent\textbf{Category 3: English Caption Benchmarks}
As an additional expansion of the ARB, we integrated two new domains—agricultural image interpretation and historical \& archaeological understanding—using visual content and captions sourced from AgriCLIP \cite{nawaz2025agriclip} and TimeTravel \cite{ghaboura2025time}, respectively. To generate Arabic reasoning questions with corresponding step-by-step answers, we adopted the \textbf{synthetic prompting} like framework inspired by \cite{shao2023synthetic} implemented using GPT-4o. This approach followed a backward–forward generation strategy; the model first synthesized a plausible reasoning chain (backward step), then generated a question that would logically yield that reasoning. In the forward step, the model refined the reasoning trace to ensure alignment and internal consistency. To ensure data quality and reasoning diversity, we applied a complexity-based selection criterion that prioritized samples involving multi-step inference or higher-order reasoning. This pipeline enabled scalable generation of rich, inference-oriented Arabic QA pairs without requiring exhaustive manual annotation.\\

\vspace{-0.75em}
\noindent\textbf{Category 4: Synthetic Data}\\
For the OCR and Document Analysis domain, we curated a set of web-sourced images containing textual content from publicly available sources~\cite{Pinterest2025}. Each image was processed using GPT-4o, which was prompted to generate Arabic QA pairs along with corresponding step-by-step reasoning. To guide the generation process, we employed a \textbf{few-shot CoT prompting} strategy, encouraging the model to produce inference-driven reasoning chains grounded in both visual and textual cues present in the images.\\

\vspace{-0.75em}
\noindent\textbf{Category 5: Tool-augmented Generated Data}\\
In this category, we constructed the domain of Charts, Diagrams, and Tables by integrating external tools to create visual samples. For the charts subdomain, data was derived from both human-curated topics and synthetic scenarios using GPT-4o under human guidance, with visualizations produced via Python and Matplotlib \cite{bisong2019matplotlib}. The tables subdomain involved generating structured data using GPT-4o and Claude-3.5 \cite{claude}, based on human-defined themes, and visualized in Excel to simulate realistic interpretation tasks. For diagrams, we adapted a subset of the AI2D dataset \cite{kembhavi2016diagram}, translating and extensively editing the content into Arabic. Human annotators refined the corresponding questions to prioritize reasoning over factual recall. Across all subdomains, GPT-4o was prompted using a \textbf{few-shot CoT} strategy to generate Arabic QA pairs with explicit step-by-step reasoning.\\

\vspace{-0.75em}
\subsection{Data Filtering and Verification Process}
To ensure the integrity and quality of ARB, we implemented a multi-stage filtering and verification pipeline (Figure~\ref{fig:data_collection_gen_verify}). This process combined manual correction, semi-automated AI–human refinement, and native speaker validation, each tailored to the complexity and origin of the data.\\

\vspace{-0.75em}
\noindent\textbf{Manual Review and Targeted Corrections:}\\
In the initial review phase, human annotators—primarily native Arabic speakers—directly corrected minor issues such as typos, grammar errors, or subtle translation inconsistencies. This approach was especially effective for Category 1, where translated content from English required adjustments rather than full regeneration. To support this workflow, we developed a custom annotation interface for efficient review (see Figure~\ref{fig:tran_verify} in Appendix~\ref{sec:app_FVI}).\\

\vspace{-0.75em}
\noindent\textbf{Iterative Human–AI Refinement:}\\
For all other categories, we adopted a semi-automated human-in-the-loop framework. GPT-4o generated step-by-step reasoning, which was then reviewed by native speakers and domain experts for logical consistency, linguistic clarity, and cultural alignment. When errors were found, such as unclear steps or reasoning gaps, the annotators provided targeted feedback, prompting partial regeneration or manual edits. This loop continued until each item met the desired quality standard. A second interface (see Figure~\ref{fig:gen_verify}, Appendix~\ref{sec:app_FVI}) allowed annotators to check, rate, flag, and finalize items efficiently.\\

\vspace{-0.25em}
\noindent\textbf{Quality Filtering and Cultural Alignment:}\\
Post-refinement, all question–answer–reasoning samples were evaluated against strict quality criteria: accuracy, coherence, reasoning completeness, and Arabic fluency. We applied both automated checks (e.g., verifying the answer aligns with the reasoning steps) and manual review. Over 200 samples were discarded at this stage due to cultural misalignment or insufficient reasoning depth. This filtering step ensured only high-quality, culturally appropriate, and challenging samples were retained.\\

\vspace{-0.75em}
\noindent\textbf{Final Approval and Integration:\\}
Items that passed all prior checks were subjected to a final review to ensure proper formatting, logical coherence, and internal consistency. Upon approval by native Arabic reviewers, the data was standardized and formally integrated into the ARB benchmark. This final validation step ensured that each entry was complete, well-structured, and suitable for robust evaluation of Arabic multimodal reasoning. Further details on the filtering, verification procedures, and annotation interfaces are provided in the Appendix ~\ref{sec:app_FVI}.

\vspace{-0.6em}

\subsection{ARB Data Statistics}
The ARB benchmark consists of 1,356 multimodal samples distributed across 11 domains (Figure~\ref{fig:domain_dist}), with Math \& Logic comprising the largest share, followed by Charts, Diagrams, \& Tables. Each sample includes an image, an Arabic question, and a set of step–action pairs leading to a final answer. In total, ARB contains 5,119 reasoning steps, with no fixed limit imposed during generation to preserve flexibility based on task complexity. Most samples include 2–6 steps, with an average of 3.78 and a median of 4. The number of steps ranges from 1 to 16, with Math \& Logic exhibiting the highest reasoning depth. Further statistics are presented in Appendix~\ref{sec:data_stat} Figure~\ref{fig:cat_step_count}.

\section{Evaluation Framework}
\subsection{Model and Prompt Selection}
\par We selected GPT-4o and GPT-4o-mini as candidate models due to their demonstrated efficiency and effectiveness in multimodal tasks, referring to ~\cite{heakl2025ain}. Recognizing the sensitivity of reasoning performance to prompt language, we evaluated both models using prompts in English and Arabic. A diverse set of 40 samples spanning multiple domains was assessed by three native Arabic speakers. To further support the evaluation of translated outputs, we employed LaBSE \cite{feng2020language} to measure semantic similarity between English and Arabic responses.

\begin{figure}[t!]
\centering  
\includegraphics[width=.48\textwidth]{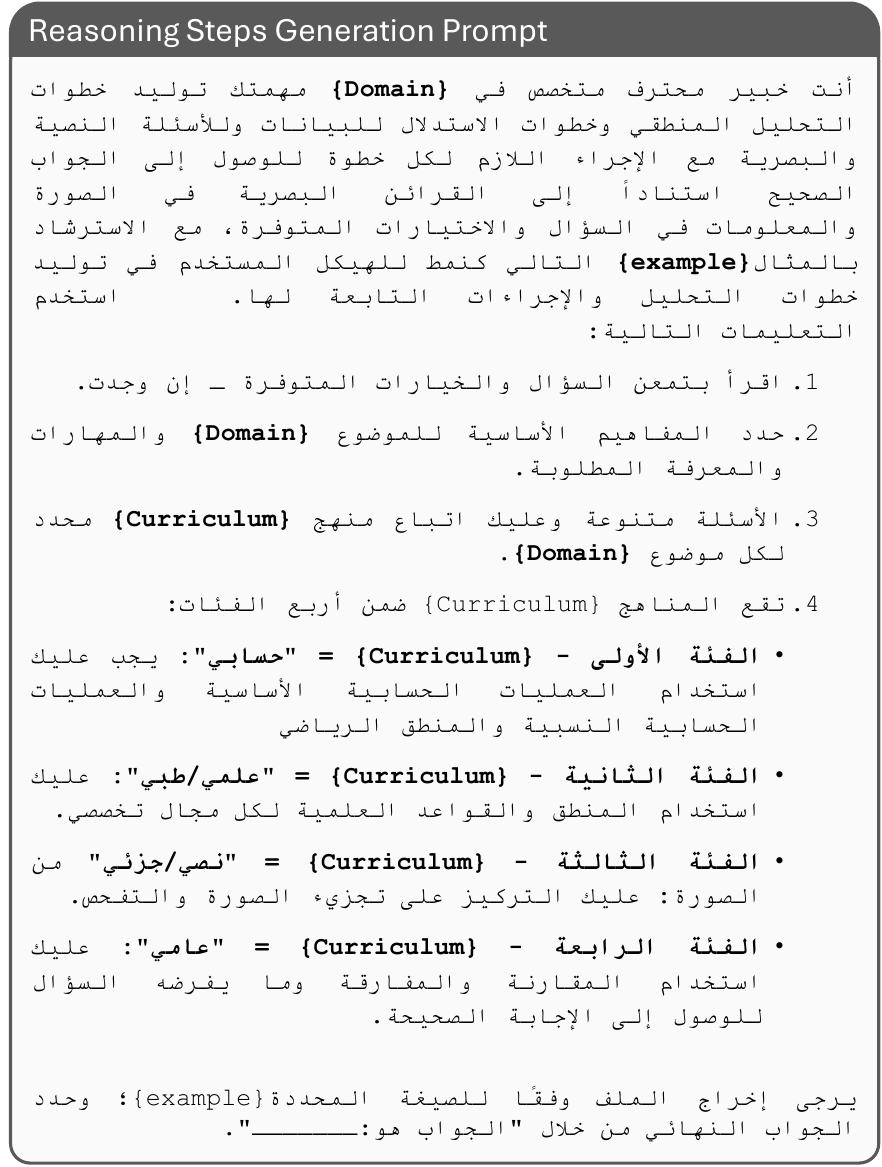}
\vspace{-1.5em}
\caption{\textbf{ARB Evaluation Prompt.}
The standardized Arabic prompt used across all ARB domains to elicit structured, curriculum-based reasoning steps from evaluated models during inference. The English version is provided in Appendix~\ref{sec:app_LLM_Prompt}.}
  \label{fig:prmpot}
  \vspace{-1.5em}
\end{figure}

Human evaluations consistently favored GPT-4o in both prompt settings. When incorporating LaBSE, GPT-4o with Arabic prompts achieved the highest similarity scores. However, across all settings, automated scores remained lower than human judgments, reflecting the models' difficulty in capturing acceptable variations in structure and order. To mitigate this, we adopted a few-shot prompting strategy, which improved similarity scores by 20–30\%, while preserving GPT-4o with Arabic prompts as the best performer. Thus, we finalize GPT-4o with Arabic prompts for the generation of reasoning steps (Figure~\ref{fig:prmpot}).

%----------- Closed Table 
\begin{table*}[!ht]
\centering
\setlength{\tabcolsep}{12pt}
\renewcommand{\arraystretch}{1}
\resizebox{0.99\textwidth}{!}{%
\begin{tabular}{lcccccc}
\toprule
\cellcolor{gray!10}{\textbf{\emph{Closed-source}}}

&\textbf{GPT-4o} 
&\textbf{GPT-4o}
&\textbf{GPT-4.1} 
&\textbf{o4}
& \textbf{Gemini 1.5}
& \textbf{Gemini 2.0}\\

\cellcolor{gray!10}{\textbf{Models}}
&&\textbf{-mini}&&\textbf{-mini}&\textbf{Pro}&\textbf{Flash}\\
\midrule
\cellcolor{gray!10}{\textbf{Final Answer (\%)}}  & \textbf{60.22} &52.22 & 59.43  &58.93& 56.70&57.80 \\
\cellcolor{gray!10}{\textbf{Reasoning Steps (\%)}} & 64.29 & 61.02 &80.41&\textbf{80.75} &64.34 & 64.09 \\

\midrule
% & & & & & & \\
\midrule

\cellcolor{gray!10}{\textbf{\emph{Open-source}}}
& \textbf{Qwen2.5}
& \textbf{Llama-3.2} 
& \textbf{AIN}
& \textbf{Llama-4}
& \textbf{Aya-}
&\textbf{InternVL3}\\
\cellcolor{gray!10}{\textbf{Model} }& \textbf{VL-7B}& \textbf{11B-Vis-Inst.} & & \textbf{Scout \small{(17Bx16E)}}&\textbf{vision-8B}& \textbf{-8B}\\
\midrule
\cellcolor{gray!10}{\textbf{Final Answer (\%)}}  & 37.02 & 25.58 & 27.35 & \textbf{48.52} & 28.81&31.04 \\
\cellcolor{gray!10}{\textbf{Reasoning Steps (\%)}} & 64.03 &53.20 & 52.77& \textbf{77.70}& 63.64& 54.50 \\  
\bottomrule
\end{tabular}}
\vspace{-0.5em}
\caption{\textbf{Stepwise Evaluation Using LLM-as-Judge.}
Comparison of closed- and open-weight models based on final answer accuracy and aggregated quality scores of reasoning steps, using our LLM-as-Judge framework with Arabic prompts and evaluation metrics. The evaluation follows a reference-based, attribute-level protocol for assessing reasoning quality. The best model in each category (closed- and open-source) is shown in bold.}

\label{tab:models_eval_all}
\end{table*}

\begin{table*}[hptb]
\centering
\setlength{\tabcolsep}{18pt}
\renewcommand{\arraystretch}{1}
\resizebox{0.99\textwidth}{!}{%
% \begin{tabular}{l|c|ccc|c|c}
\begin{tabular}{cl|c|ccc|c|c}
\toprule
&\cellcolor{gray!10}{\textbf{Model}} &\textbf{BLEU} &\textbf{ ROUGE-1} & \textbf{ROUGE-2 }& \textbf{ROUGE-L}& \textbf{BERTScore} &\textbf{LaBSE}\\
\midrule
\multirow{6}{*}{\rotatebox[origin=c]{90}{Closed-source}} 

&\cellcolor{gray!10}{GPT-4o} & 6.21 & 63.61& 42.71 &58.70&76.33 &82.82\\

&\cellcolor{gray!10}{GPT-4o-mini}& 5.30 & 61.86& 41.18 &56.73& 76.23 &81.56\\

&\cellcolor{gray!10}{GPT-4.1} & 6.35 &\textbf{71.13} &48.83 & 65.33& 77.32& \textbf{84.40}\\

&\cellcolor{gray!10}{o4-mini}& 5.38 & 65.22 & 45.94 & 59.45& 76.33 & 82.57\\

&\cellcolor{gray!10}{Gemini 1.5 Pro} & 5.49 & 62.71& 45.90&58.34& 76.05 &79.81\\

&\cellcolor{gray!10}{Gemini 2.0 Flash}& \textbf{8.27 }& 70.91 &\textbf{ 54.81} &\textbf{ 65.95} & \textbf{78.56 }&83.77 \\

\midrule

\multirow{6}{*}{\rotatebox[origin=c]{90}{Open-source}}

&\cellcolor{gray!10}{Qwen2.5-VL-7B}& 3.21 &48.51  & 31.19 &  45.97& 73.03 &73.67\\

&\cellcolor{gray!10}{Llama-3.2-11B }& 1.75 & 22.83 & 11.20 & 19.63 & 66.89  &65.41\\

&\cellcolor{gray!10}{AIN} & 3.16 & 59.18 & \textbf{43.54} & \textbf{55.41 }&\textbf{73.26} &72.25\\

&\cellcolor{gray!10}{Llama-4 Scout} &\textbf{4.32} & 47.74 & 27.52 & 41.07 & 73.06 &\textbf{77.51}\\

&\cellcolor{gray!10}{Aya-Vision-8B} &3.39 & \textbf{59.64} & 38.98 & 53.80 & 72.54 &76.84\\

&\cellcolor{gray!10}{IntenVL3-8B} &2.93 & 50.78 &29.96  & 46.35 &  72.52 &77.28\\
\bottomrule
\end{tabular}}
\vspace{-0.75em}
\caption{\textbf{Lexical and Semantic Similarity Scores.} Evaluation of generated reasoning steps using classical metrics, including BLEU, ROUGE, BERTScore, and LaBSE. These metrics reflect surface-level lexical overlap and overall semantic similarity but fall short in capturing stepwise logical coherence. The best model in each category (closed- and open-source) is shown in bold.}
\vspace{-0.5em}
\label{tab:sementic}
\end{table*}

\subsection{Evaluation Methodology and Metrics}
\textbf{Lexical and Semantic Similarity Metrics.} 
\par To assess similarity between predicted reasoning steps and human-curated references, we employed standard metrics (Table~\ref{tab:sementic}). BLEU~\cite{papineni2002bleu} showed weak n-gram alignment, while ROUGE variants~\cite{lin2004rouge} yielded mixed results with a sharp drop in ROUGE-2, indicating limited fluency. For semantic similarity, we used BERTScore~\cite{zhang2019bertscore}, which captures token-level alignment but lacks cross-lingual robustness, reducing its reliability for Arabic evaluation. To address this, we adopted LaBSE~\cite{feng2020language}, a multilingual sentence-level model that provided more stable results, averaging 81.5\%±2 for closed-weight models and 71.5\%±5 for open-weight ones. Despite their utility, these metrics fall short in capturing logical structure, coherence, and factual grounding in multi-step reasoning.\\

\label{sec:judge_prompt}
\begin{figure*}[t!]
\centering  
\includegraphics[width=\textwidth,height=5.5cm]{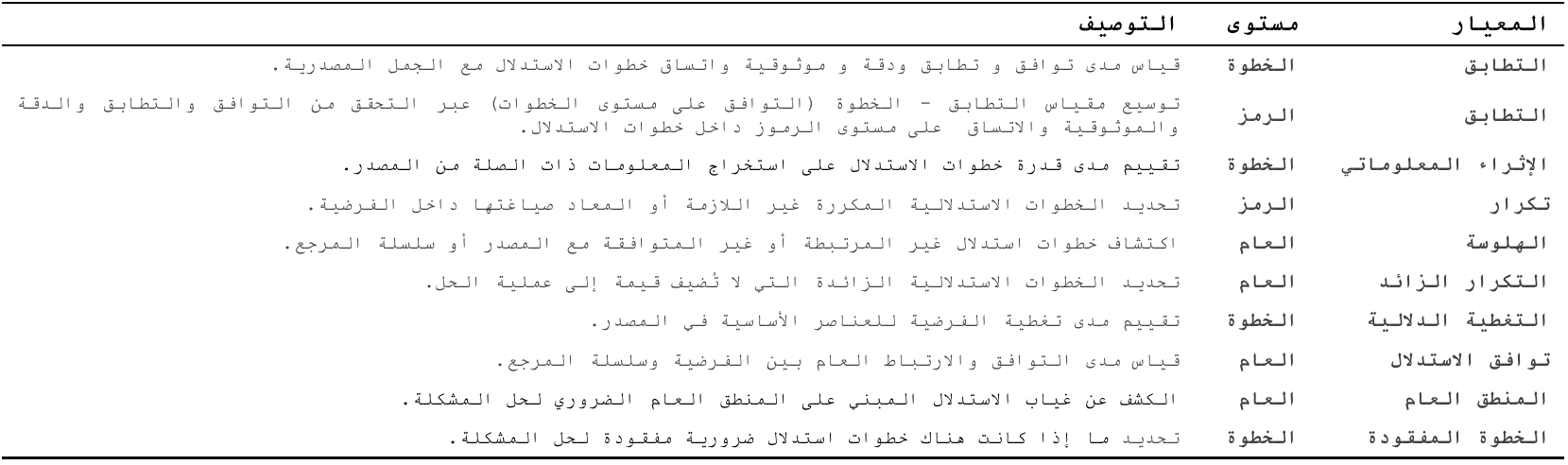}
\vspace{-1.75em}
\caption{\textbf{Arabic Reasoning Evaluation Metrics.}
We assess step-by-step reasoning using five core Arabic-specific dimensions: \emph{Faithfulness (At-Tatābuq)}, \emph{Informativeness (Al-Ithrā' Al-Ma'lūmātī)}, \emph{Coherence (At-Tawāfuq)}, \emph{Commonsense (Al-Mantiq Al-'Āmm)}, and \emph{Reasoning Alignment (At-Tawāfuq Al-Istidlālī)}. Auxiliary checks cover hallucinations, redundancy, semantic gaps, and missing steps. Metrics are defined at the step and/or token level. The full evaluation rubric is provided in English in Appendix~\ref{sec:app_LLM_Prompt}.}

\label{fig:metrics}
\vspace{-1.25em}
\end{figure*}  
\noindent\textbf{Stepwise Evaluation Using LLM-as-Judge}
\par To address the limitations of traditional evaluation metrics, we adopted a structured LLM-as-Judge framework, along with a reference-based protocol and Arabic prompt, adapted from \cite{thawakar2025llamav} evaluation suite. Unlike reference-free metrics \cite{golovneva2022roscoe}, this set-up enables a fine-grained, interpretable evaluation aligned with Arabic linguistic and contextual nuances. GPT-4o, used as LLM-as-Judge, is instructed to assess reasoning outputs across several dimensions, including faithfulness, informativeness, redundancy, hallucination, semantic coverage, and commonsense reasoning. Each attribute is rated on a scale from 1 to 10 (see Figure~\ref{fig:heatmap_closed} and Figure~\ref{fig:heatmap_open}), and the final score for reasoning steps is computed as the average across all dimensions (Table~\ref{tab:models_eval_all}). The full evaluation prompt is provided in Appendix~\ref{sec:app_eval_prompt}.

% \vspace{-0.7em}
\noindent\textbf{Inter-Annotator Agreement: Krippendorff’s Alpha.} To ensure data quality and validate the efficiency of our LLM-as-Judge selection, we conducted an inter-annotator agreement analysis over 5\% of the dataset. Three human annotators were provided with a user-friendly interface (Figure~\ref{fig:IAA_GUI}) to rate samples on a scale from 1 (lowest) to 5 (highest). Most samples received scores of 4 or higher, confirming the effectiveness of our earlier verification steps and reflecting strong agreement among annotators. We measured Krippendorff’s Alpha \cite{krippendorff2018content}, achieving a score of 83.56\% among human annotators. To further assess the reliability of GPT-4o as an LLM-as-Judge, we repeated the evaluation by including the model's judgments, resulting in an even higher Krippendorff’s Alpha of 87.62\%. These results demonstrate high consistency between human and LLM assessments, supporting the robustness of our evaluation framework.

\begin{figure*}[hptb]
\centering  
\includegraphics[width=\textwidth]{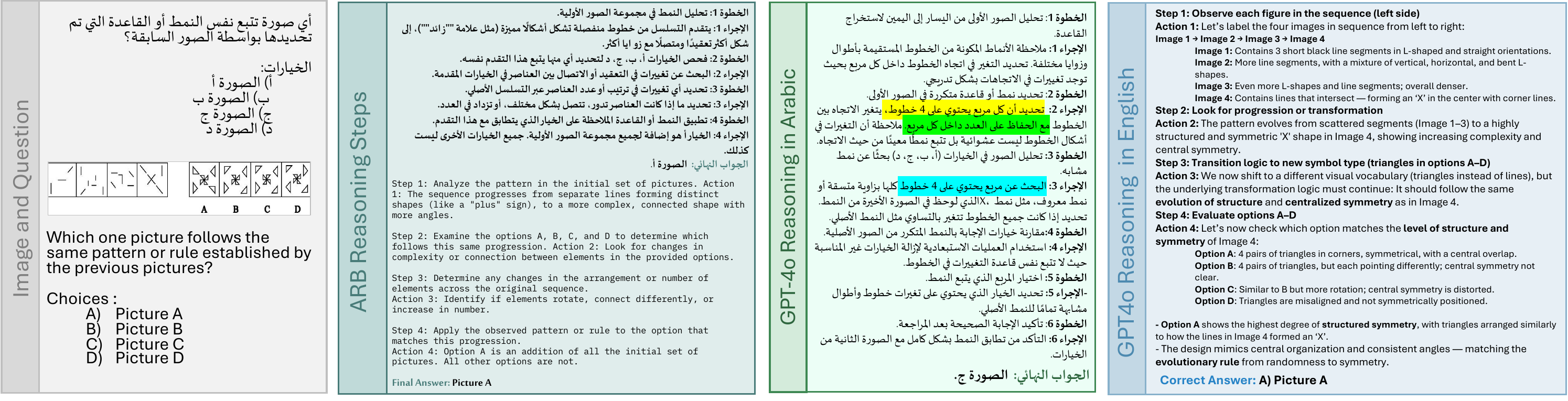}
\vspace{-1.5em}
\caption{\textbf{Cross-Lingual Reasoning Comparison (Arabic vs. English).} 
This figure compares LMMs (GPT-4o) reasoning steps in Arabic and English for the same visual task. In the Arabic version, the model misinterprets structural constraints, \colorbox{yellow}{yellow} highlights incorrect assumptions about equal line counts across boxes, \colorbox{green}{green} emphasizes miscounted lines within the boxes, and \colorbox{highlightcyan}{cyan} marks an irrelevant search for a box with exactly 4 lines. These reasoning flaws lead to the wrong answer (C). In contrast, the English reasoning is structured, accurate, and constraint-aware, correctly identifying the answer (A), highlighting the performance gap in Arabic.}
\label{fig:ar-en-eval}
\vspace{-1.25em}
\end{figure*}  

\section{Results and Analysis}
\label{sec:res_analysis}
\textbf{Reasoning–Answer Performance Gap.}
\par The ARB evaluation (Table~\ref{tab:models_eval_all}) reveals a consistent gap between models' ability to generate coherent reasoning steps and their success in reaching correct final answers. For example, models like GPT-4.1 and o4-mini achieve reasoning coherence scores above 80\%, while their final answer accuracy hovers around 58–60\%. This pattern is even more pronounced in open models such as Qwen2.5-VL and Aya-vision, where reasoning steps are moderately strong (above 50–60\%) but final answer correctness remains below 40\%. These results demonstrate that well-structured reasoning does not guarantee correct conclusions—underscoring the need for step-level evaluation to accurately assess a model’s reasoning capabilities.\\

\vspace{-0.75em}
\noindent\textbf{Closed vs. Open-Source Model Performance.}
\par \textbf{Quantitative Evaluation.} Closed-source models consistently outperform open-source ones in both reasoning and final answer accuracy. GPT-4.1 and o4-mini lead the closed category, with strong logical consistency and relatively high correctness. Among open models, LLaMA-4 Scout performs best, scoring 77.7\% in reasoning steps and 48.5\% in final answers—narrowing the gap with closed models but still trailing. Other open models such as LLaMA-3.2, AIN, Aya Vision, and InternVL3 demonstrate coherent reasoning but struggle with accurate conclusions, reflecting limitations in cross-lingual understanding and cultural grounding.\\

\vspace{-0.5em}

\par \textbf{Qualitative Evaluation.} 
To investigate reasoning gaps in Arabic, we conducted a qualitative comparison between model outputs and human-curated ARB references. Selected examples illustrate common pitfalls in both open- and closed-source models, including incomplete or incoherent step transitions, hallucinations, and shallow logical progression in Arabic responses (Figures~\ref{fig:open_qual} and \ref{fig:closed_qual}).

We further examine the impact of language by comparing Arabic and English reasoning steps generated by the same model on identical visual inputs (Figure~\ref{fig:ar-en-eval}). This side-by-side analysis reveals notable inconsistencies in reasoning quality across languages, emphasizing the need for Arabic-specific benchmarks.

These findings underscore the importance of evaluating and improving Arabic multimodal reasoning, directly supporting ARB’s core motivation.\\

\par\textbf{Domain-Level Trends.}
Figures~\ref{fig:hcharts_closed} and~\ref{fig:hcharts_open} (Appendix~\ref{sec:app_Charts}) show a domain-level breakdown, illustrating the persistent reasoning-answer gap across task categories. Figures~\ref{fig:heatmap_closed} and~\ref{fig:heatmap_open} offer fine-grained step-by-step scores, revealing domain-specific model behavior. These results underscore \textsc{ARB}’s value in exposing nuanced reasoning patterns and highlighting the strengths and weaknesses of both closed- and open-source models across domains.

\vspace{-0.5em}
\section{Conclusion}
In this work, we presented ARB, the first benchmark designed to evaluate step-by-step multimodal reasoning in Arabic across 11 diverse domains. With 1.35K high-quality samples and over 5K human-curated reasoning steps, it was built through a hybrid pipeline combining prompting strategies, tool-assisted generation, and native-speaker validation. Our evaluation of 12 state-of-the-art open- and closed-weight models revealed persistent gaps in reasoning quality, coherence, and cultural alignment when operating in Arabic, despite their strong performance in English-centric settings. These findings underscore the need for step-level, culturally aware evaluation strategies tailored to underrepresented languages. Beyond benchmarking, the open-source ARB offers tools, protocols, and interfaces to support reproducibility and future research. It sets the foundation for training and evaluating Arabic-native LMMs and contributes toward building more inclusive, interpretable, and linguistically grounded AI systems.

\section{Limitations and Societal Impact}

\par While ARB provides a valuable resource for evaluating Arabic multimodal reasoning, it has certain limitations. First, although it spans 11 diverse domains, the benchmark may still not fully capture the full linguistic, dialectal, or cultural variability present across the Arabic-speaking world. Additionally, reasoning evaluations rely on human judgment and model-specific prompts, which may introduce subjectivity or prompt-induced biases. The benchmark also focuses on Arabic exclusively, and does not offer multilingual alignment or cross-lingual transfer assessments, which could be valuable for comparative studies.

\par From a societal perspective, ARB promotes more inclusive and culturally aware AI by centering Arabic, an underrepresented yet widely spoken language. Its focus on interpretable, step-by-step reasoning supports broader goals of AI transparency and accountability. Nonetheless, ethical considerations remain important, particularly to prevent the misuse or misinterpretation of culturally sensitive content in applications where AI decisions may have real-world consequences.

\bibliography{arxiv}

\clearpage
\appendix
\section{Appendix}
\label{sec:appendix}

This appendix provides supplementary material supporting our contributions. It includes: (1) a brief overview of related work situating our approach within broader research on Arabic reasoning and multimodal data generation; (2) details of the filtering and verification pipeline, including interface designs used for human-in-the-loop validation and the inter-annotator agreement study; (3) additional details on the prompts used for model reasoning generation and evaluation; (4) English translations of the Arabic generation prompt and evaluation metrics; and (5) extended data statistics, such as domain and steps by domain distributions, token length distributions in questions and reasoning steps, as well as their ratios. These additions enhance transparency and offer deeper insight into the construction and quality control of the ARB benchmark.

\section{Related Work}
\label{sec:app_RW}
\textbf{Chain-of-Thought Reasoning in LLMs}

CoT prompting was introduced by \cite{wei2022chain} to improve LLMs' logical reasoning, inspiring extensions like self-consistency \cite{wang2022self}, tree-of-thoughts \cite{yao2023tree}, and instruction tuning for reasoning \cite{vaillancourt2024instruction, ranaldi2024self}. Recent work has also explored structural aspects of reasoning, including the impact of step length \cite{jin2024impact} and counterfactual prompting to reduce bias \cite{moore2024reasoning}.

Building on these developments, state-of-the-art LLMs have adopted advanced post-training strategies to strengthen reasoning. Kumar et al. \cite{kumar2025llm} survey techniques such as fine-tuning, reinforcement learning, and test-time scaling. OpenAI’s o1 model \cite{jaech2024openai} leverages reinforcement learning and inference-time scaling to improve reasoning fidelity. DeepSeek R1 \cite{guo2025deepseek} enhances CoT performance using reward models that prioritize logical soundness over natural phrasing.\\

\vspace{-0.5em}
\noindent\textbf{Multimodal Reasoning in VLMs}

Extending CoT reasoning to multimodal tasks has proven both challenging and rewarding. Models like LLaVA-CoT \cite{xu2025llavacotletvisionlanguage} explicitly incorporate structured visual reasoning steps into their outputs, enabling multi-stage perception and interpretation of images. Trained on a dataset of 100k CoT-annotated visual QA pairs, LLaVA-CoT achieves notable gains on reasoning benchmarks. Similarly, LlamaV-o1 \cite{thawakar2025llamav} introduces a curriculum-based framework and benchmark for multi-step visual reasoning, demonstrating improvements in both accuracy and interpretability.

Recent studies have proposed methods to further enhance reasoning coherence and alignment. Chen et al. \cite{chen2024measuring} present metrics and a two-stage training strategy to improve consistency in vision-language reasoning. Zhang et al. \cite{zhang2024improve} enrich training data with rationales distilled from GPT-4o and apply Direct Preference Optimization (DPO) to guide models toward more faithful and coherent CoT outputs.

These developments reflect a growing consensus that multimodal models must reason systematically across modalities—not merely generate final answers—to ensure robustness and interoperability.\\
\vspace{-0.5em}

\noindent\textbf{Arabic and Multilingual Reasoning Resources}

Despite increasing multilingual training in LLMs, Arabic remains underrepresented in reasoning-focused benchmarks. Several datasets have emerged to address this gap. ArabicSense \cite{lamsiyah2025arabicsense} evaluates commonsense reasoning in Arabic, while AraSTEM \cite{mustapha2024arastem} offers over 11,000 science-focused multiple-choice questions in Arabic. ArabLegalEval \cite{hijazi2024arablegaleval} benchmarks Arabic legal reasoning using real-world legal documents and synthetic questions. ArabCulture \cite{sadallah2025commonsense} focuses on MSA commonsense reasoning across 13 Arab countries using culturally grounded, native-authored questions. AraDiCE \cite{mousi2024aradice} evaluates dialectal and cultural reasoning across Arabic varieties using post-edited synthetic data.

These resources reveal substantial performance disparities between Arabic and English, particularly in reasoning-heavy tasks; however, they remain limited to the text modality and focus primarily on LLMs rather than LMMs.\\

\vspace{-0.5em}
\noindent\textbf{Arabic-Native Reasoning Models}

Recent efforts have introduced Arabic-native LLMs with enhanced reasoning capabilities. ALLaM-Thinking \cite{almaghrabima2025allam} is a fine-tuned model specifically optimized for stepwise logic and arithmetic problem-solving, demonstrating improved chain-of-thought performance in math tasks through Unsloth and Grouped Policy Optimization. Fanar \cite{team2025fanar}, a broader Arabic LLM, recently introduced the ``Think Before Responding'' feature, enabling intermediate reasoning traces during decoding and improving interpretability and alignment with structured reasoning. In contrast, models like AIN \cite{heakl2025ain} and Jais \cite{sengupta2023jais} offer general Arabic capabilities but lack fine-grained reasoning alignment.\\

ARB complements these resources by providing the first multimodal step-by-step reasoning benchmark in Arabic, creating a unified framework for evaluating reasoning transparency across vision-language tasks.

\section{Filtering and Verification Pipeline and Interface}
\label{sec:app_FVI}
To ensure quality and consistency across all samples, we developed a streamlined and user-friendly annotation interface to support manual verification and scoring. Given the scale of data and multiple annotators involved, the interface was designed to simplify inspection and accelerate review.

\begin{figure}[t]
\centering

\begin{subfigure}[t]{0.5\textwidth}
    \includegraphics[width=\textwidth]{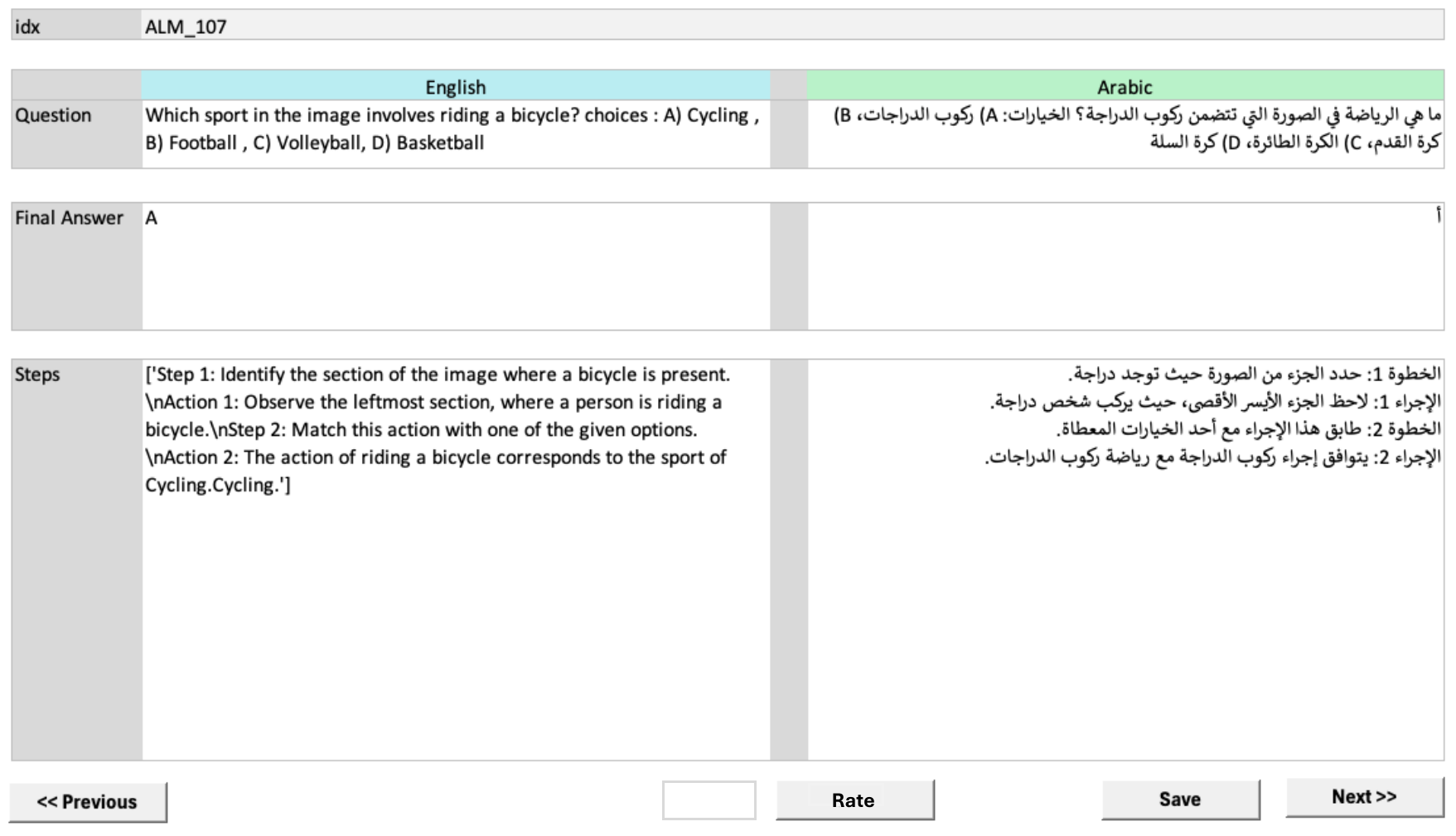}
    \caption{Example of ARB translation verification user interface.}
    \label{fig:tran_verify}
\end{subfigure}
\\
\vspace{1em}
\begin{subfigure}[t]{0.5\textwidth}
    \includegraphics[width=\textwidth]{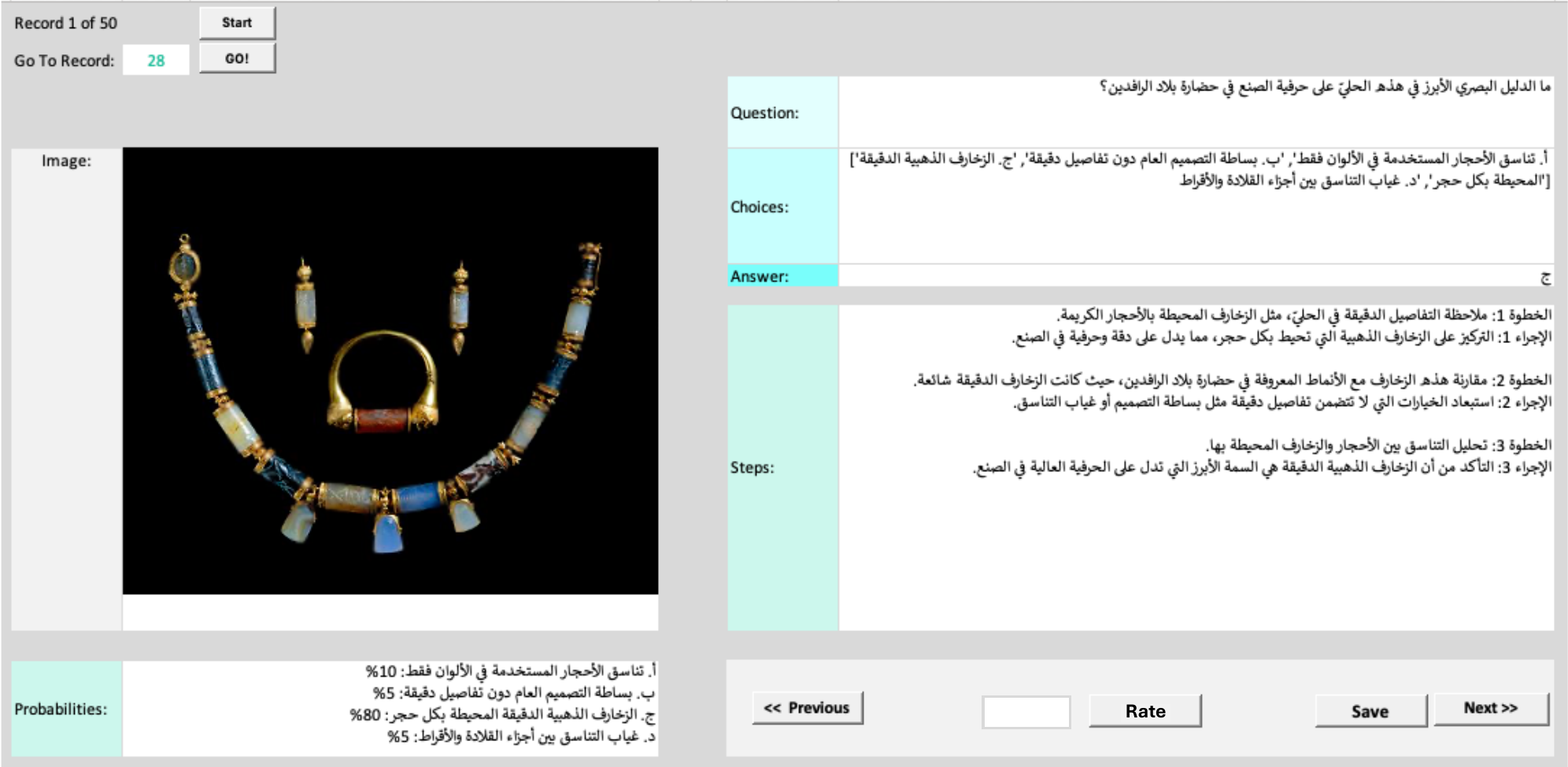} 
    \caption{Example of ARB generated data verification user interface.}
    \label{fig:gen_verify}
\end{subfigure}

\vspace{-0.5em}
\caption{\textbf{Filtering and Verification User Interface.} The interface enables annotators to manually edit content when applicable and assign quality ratings to guide subsequent controller review and final approval.}
\label{fig:verify_interfaces}
\end{figure}

For translation tasks (see Figure~\ref{fig:tran_verify}), the interface displays the original English text alongside the Arabic translation, allowing annotators to directly edit only the translated portion. For synthetic samples (see Figure~\ref{fig:gen_verify}), the interface presents the image, Arabic question, step-by-step reasoning, predicted answer, and reference answer. Annotators assess the sample based on accuracy, clarity, cultural alignment, and faithful delivery of meaning, with an emphasis on conceptual correctness rather than word-for-word translation.

Each sample is rated on a 6-point scale, as shown below.

\begin{table}[h]
\centering
\small
\begin{tabular}{cl}
\textbf{Rate} & \textbf{Description} \\
\midrule
0 & \textcolor{ppl}{Reject:} Culturally inappropriate/ Irrelevant content \\
1 & \textcolor{pinky}{Reject:} Requires full regeneration by the model \\
2 & \textcolor{yel}{Poor:} Major edits needed to fix reasoning or clarity \\
3 & \textcolor{SkyBlue}{Fair:} Moderate edits required \\
4 & \textcolor{grn}{Good:} Minor edits needed \\
5 & \textcolor{Teal}{Excellent:} No edits needed; ready for inclusion \\
\bottomrule
\end{tabular}
\caption{\textbf{Filtering and Verification Rating Scale.} A standardized scoring scheme used by annotators to assess the quality of translations and reasoning steps. The scale guides decisions on whether a sample should be accepted, revised, or regenerated based on linguistic accuracy, reasoning clarity, and cultural appropriateness.}
\vspace{-1em}
\label{tab:rating_scale}
\end{table}

Each sample was independently reviewed by two annotators and then passed to a controller, with individual scores combined for a total of 10. If either annotator assigned a score of 0, the sample was immediately discarded due to cultural or contextual inappropriateness. Samples scoring 8–10 were approved without further review, while those scoring 2–4 were sent back for regeneration. Samples with intermediate scores (5–7) were escalated to a controller, who conducted a final review, resolved discrepancies, and made any necessary corrections. This multi-tiered evaluation process ensured both the consistency and quality of the final dataset.

\vspace{-2em}

\begin{figure}[t!]
\centering  
\includegraphics[width=0.5\textwidth]{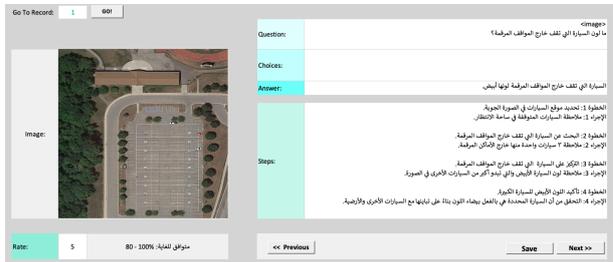}
\caption{\textbf{Inter-Annotator Agreement Interface.} The interface allows annotators to evaluate each sample by assessing the compatibility of the model's step/action chain with the provided image, question, and choices (when applicable). Annotators assign a score by comparing the model's reasoning process to their own human reasoning approach for solving the question.}
\label{fig:IAA_GUI}
\vspace{-1em}
\end{figure}  

% \vspace{-2em}
\section{Models' Evaluation Prompts}
\label{sec:app_eval_prompt}

\par
\begin{figure}[t!]
\centering  
\includegraphics[width=0.48\textwidth,height=12cm]{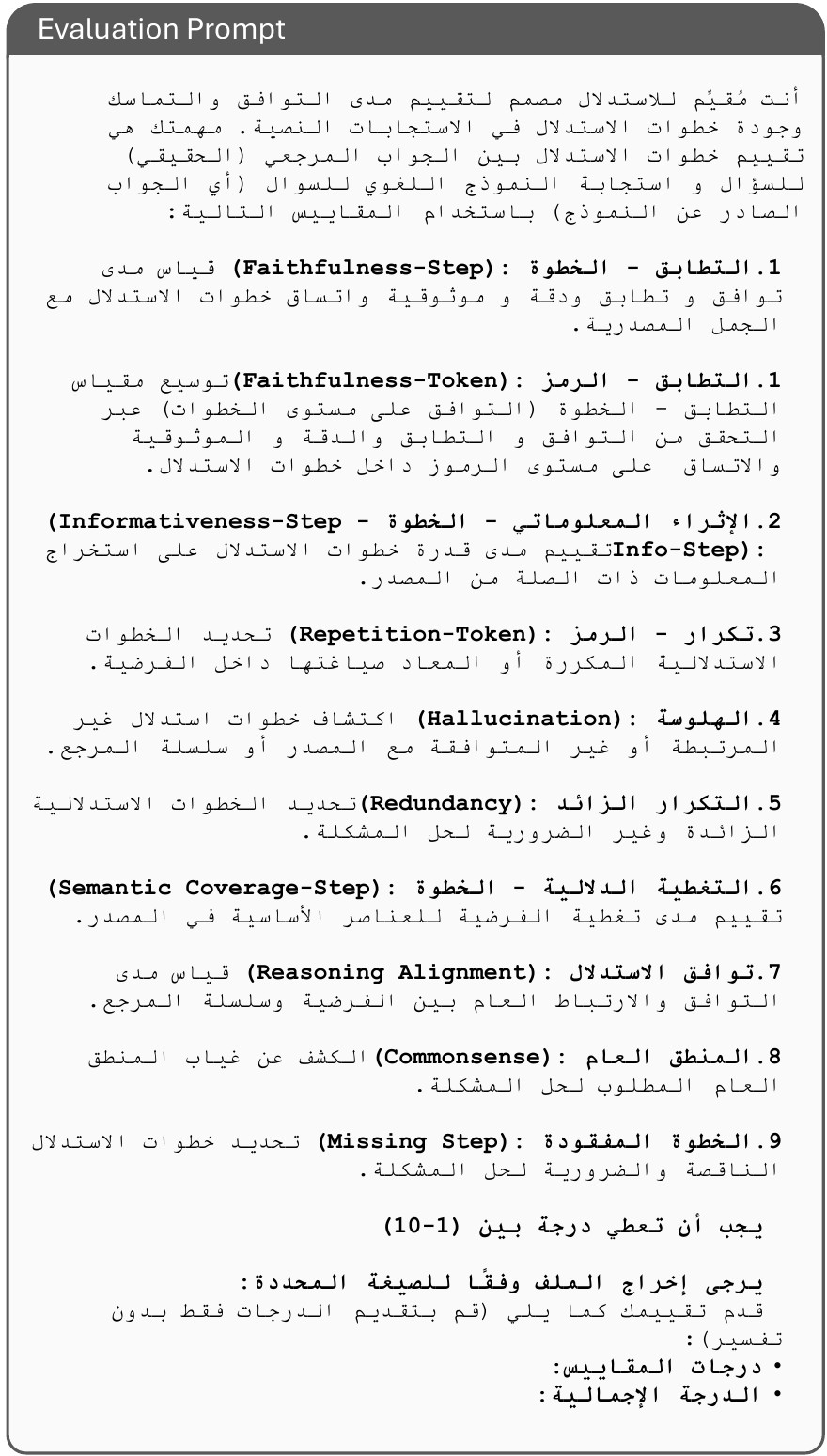}
\vspace{-1.5em}
\caption{\textbf{Arabic Evaluation Prompt for LLM-as-Judge.} 
This prompt was used to evaluate reasoning steps across all models in Arabic. It guides models to assess reasoning quality using a set of structured criteria defined in the ARB framework.}
\label{fig:eval_prompt_ar}
\vspace{-1.5em}
\end{figure}

 This section presents the evaluation prompts used to assess the step-by-step reasoning quality of LMMs in our study. The prompt was adapted from the LLamaV-o1 evaluation protocol \cite{thawakar2025llamav} and tailored to the Arabic multimodal reasoning context of ARB (Figure~\ref{fig:eval_prompt_ar}). To ensure consistency between the generation and evaluation phases, all assessments were performed using Arabic prompts exclusively in open-source and closed-source models. This design choice maintained linguistic alignment with model outputs and minimized potential cross-lingual biases during judgment.

\begin{figure}[t!]
\centering  
\includegraphics[width=0.48\textwidth,height=11cm]{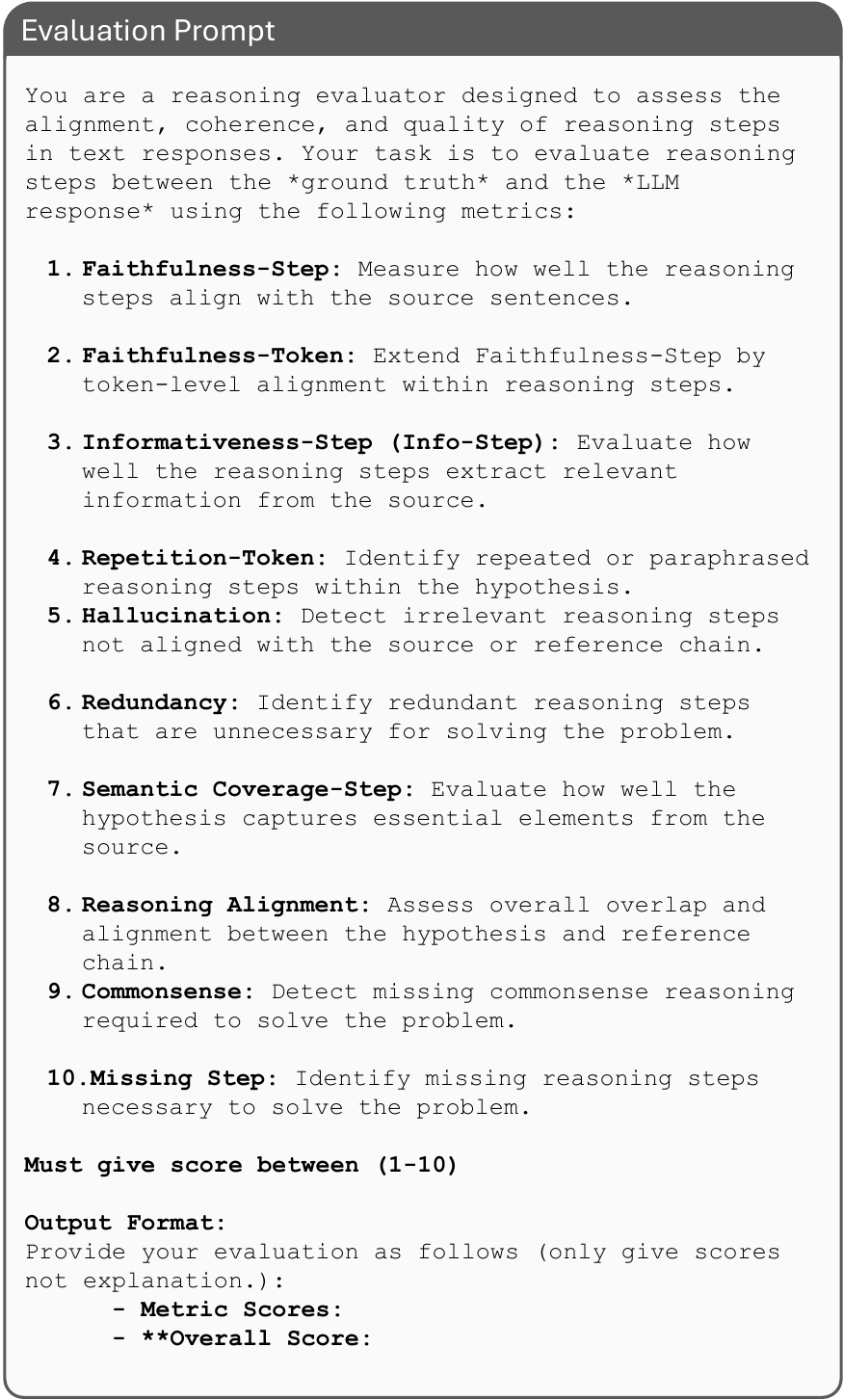}
\vspace{-1.5em}
\caption{\textbf{English Translation of the Arabic Evaluation Prompt.} 
A translated version of the prompt used to evaluate reasoning steps in ARB (see Figure~\ref{fig:eval_prompt_ar}) to aid non-Arabic readers.}
\label{fig:eval_prompt_en}
\vspace{-1em}
\end{figure}

An English translation of the prompt is provided (Figure~\ref{fig:eval_prompt_en}) to assist non-Arabic readers and enhance accessibility.

\section{English Translation of Generation Prompt and Evaluation Metrics}
\label{sec:app_LLM_Prompt}

\par This section presents the English translations of two core components used in ARB: (1) the prompt for the generation of reasoning steps, originally designed in Arabic (see the Arabic version in Figure~\ref{fig:prmpot}, the English translation in Figure~\ref{fig:prompt_en}); and (2) the evaluation metrics used to assess the quality of these reasoning steps (see original in Figure~\ref{fig:metrics}, the English translation in Figure~\ref{fig:metrics_en}). These metrics were also used in the evaluation prompt provided in Appendix~\ref{sec:app_eval_prompt}.

\begin{figure*}[t!]
\centering  
\includegraphics[width=\textwidth,height=9cm]{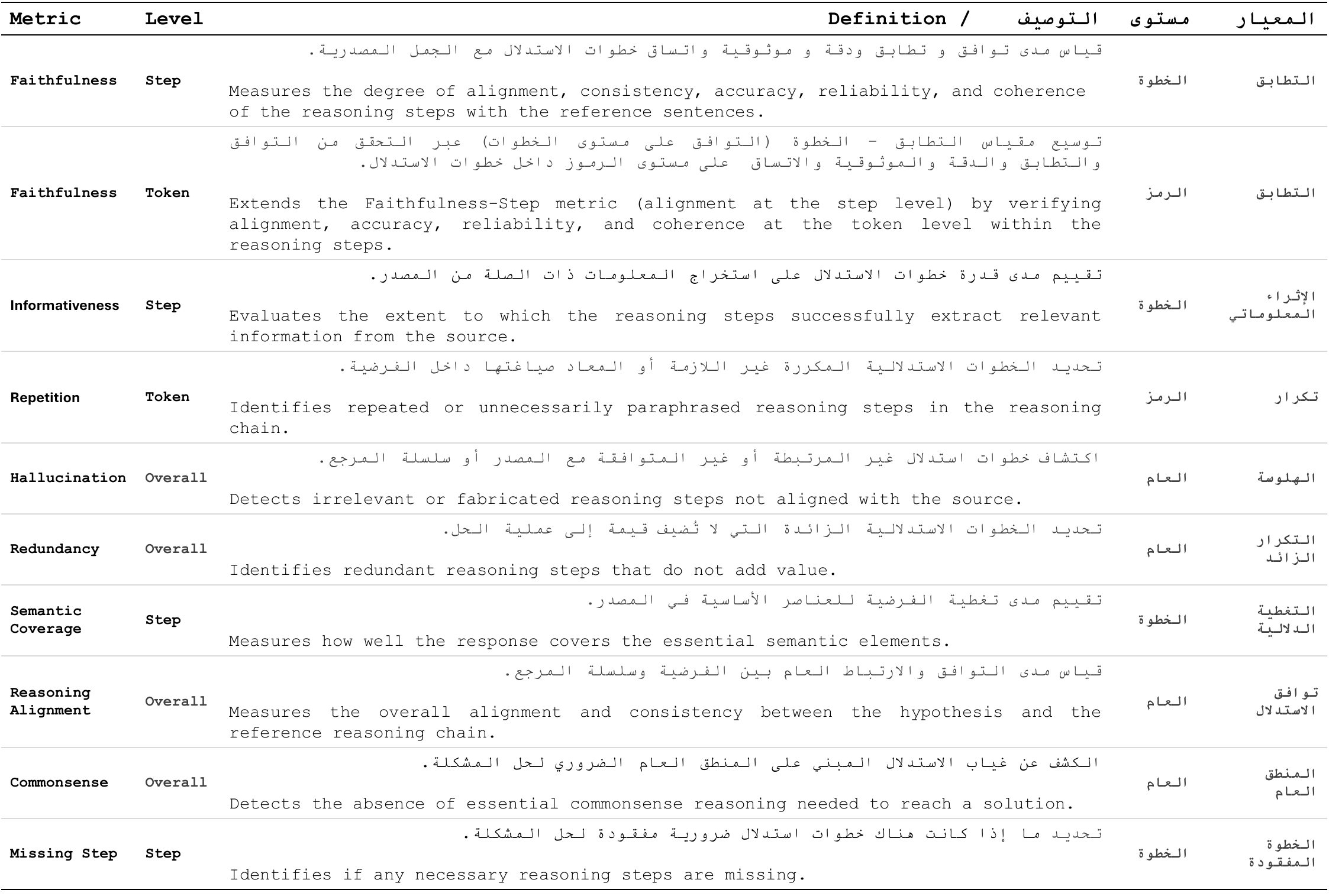}
\vspace{-1.5em}
\caption{\textbf{English Translation of ARB Evaluation Metrics.}  
An English version of the Arabic reasoning evaluation rubric used in ARB (see Figure~\ref{fig:metrics}), detailing the definitions of all step-level and overall reasoning quality metrics. These include measures for faithfulness, informativeness, repetition, hallucination, redundancy, semantic coverage, reasoning alignment, commonsense reasoning, and missing steps. This translation supports cross-lingual reproducibility and interpretability of the evaluation framework.}
\label{fig:metrics_en}
\vspace{-1.25em}
\end{figure*}  

\begin{figure}[hptb]
\centering  
\includegraphics[width=.48\textwidth]{figures/prompt_en.pdf}
\vspace{-1.5em}
\caption{\textbf{English Version of the ARB Prompt.}  
This figure presents the English translation of the original Arabic prompt (see Figure~\ref{fig:prmpot}) used to guide reasoning step generation across domains.}
  \label{fig:prompt_en}
  \vspace{-1.25em}
\end{figure}

\section{Domain-Level Analysis of Reasoning and Final Answers}
\label{sec:app_Charts}
\par To gain deeper insight into model performance across various task categories, we present a domain-level analysis of ARB results for both closed- and open-source models. These visualizations illustrate how models perform in terms of both final answer accuracy and reasoning step quality across the 11 benchmark domains. \\
To support clarity and consistency across the following visual analyses, we adopt the following standardized abbreviations for the 11 ARB domains:

\begin{table}[hptb]
    \centering
    \vspace{-1em}
    \begin{tabular}{ll}
    \textbf{Abb} & \textbf{Description} \\
\midrule
\small \textbf{VR}& \small Visual Reasoning;\\
\small \textbf{OCR}& \small OCR and Document Analysis;\\
\small \textbf{CDT}&  \small Charts, Diagrams, and Tables;\\
\small \textbf{M\&L}&\small Mathematical and Logical Reasoning;\\
\small \textbf{Soc.Cult.}& \small Social and Cultural Understanding;\\
\small \textbf{CVP}& \small Complex Visual Perception;\\
\small \textbf{MED}&\small Medical Image Analysis;\\
\small \textbf{Sci.R}&\small Scientific Reasoning;\\
\small \textbf{Hist.}&\small Historical \& Archaeological Interpretation;\\
\small \textbf{RS}& \small Remote Sensing Analysis;\\
\small \textbf{Agro}&\small  Agricultural Image Understanding.
    \end{tabular}
    \label{tab:abbrev}
\end{table}

\begin{figure*}[hptb]
\centering  
\includegraphics[width=\textwidth,height=10cm]{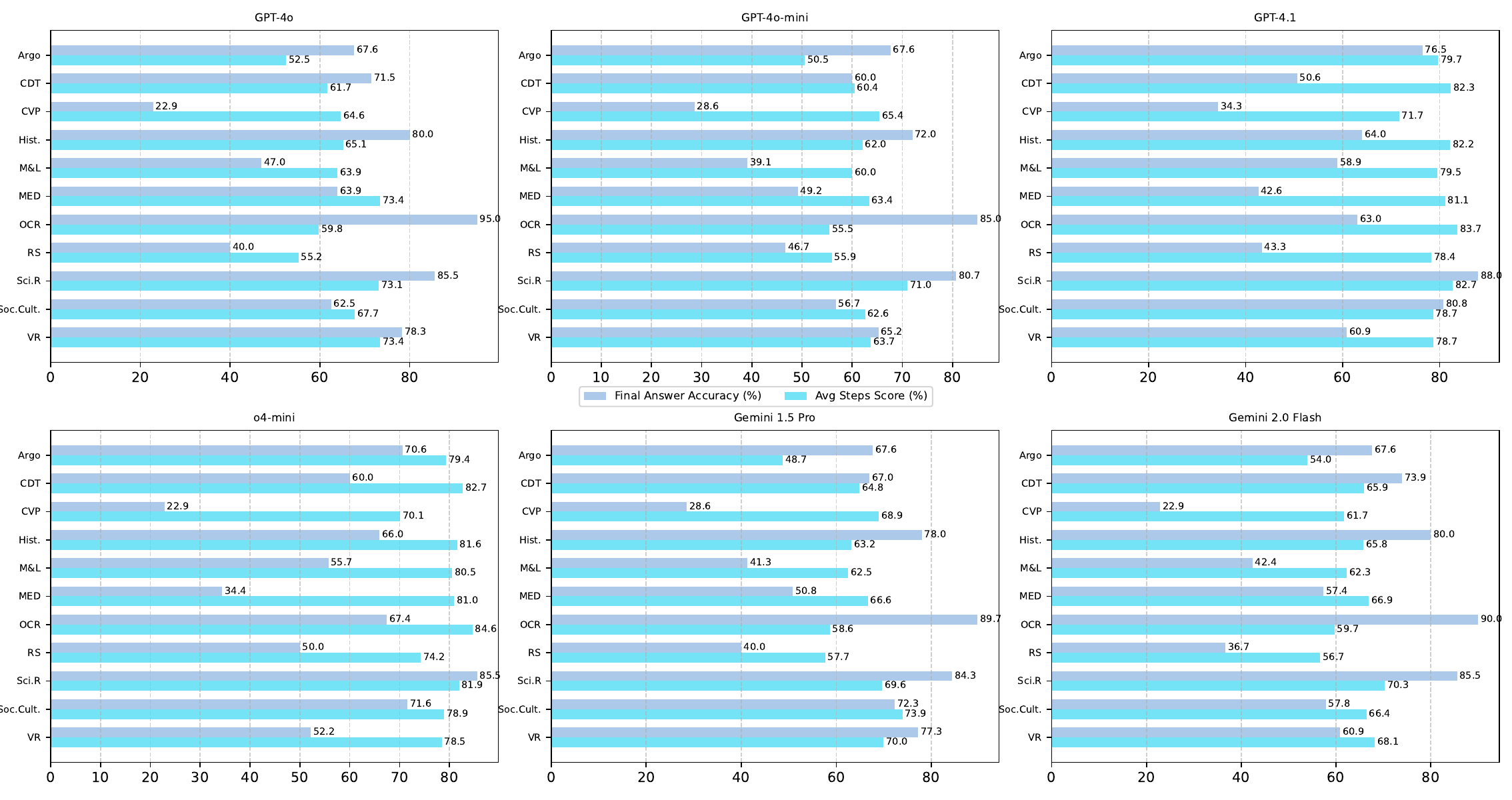}
\vspace{-1.5em}
  \caption{\textbf{Domain-Level Performance of Closed-Source Models.}  
  Bar charts comparing final answer accuracy and average reasoning step quality across ARB domains for each closed-source model. GPT-4.1 and o4-mini show strong reasoning in domains like Sci.R, CDT, and Hist., while notable gaps appear in CVP and RS. All models consistently score higher on reasoning than final answers, underscoring the importance of step-level evaluation. The figure highlights both strengths and limits of closed models in Arabic multimodal reasoning.}

  \label{fig:hcharts_closed}
\vspace{-1em}
\end{figure*}

\begin{figure*}[hptb]
\centering  
\includegraphics[width=\textwidth,height=10cm]{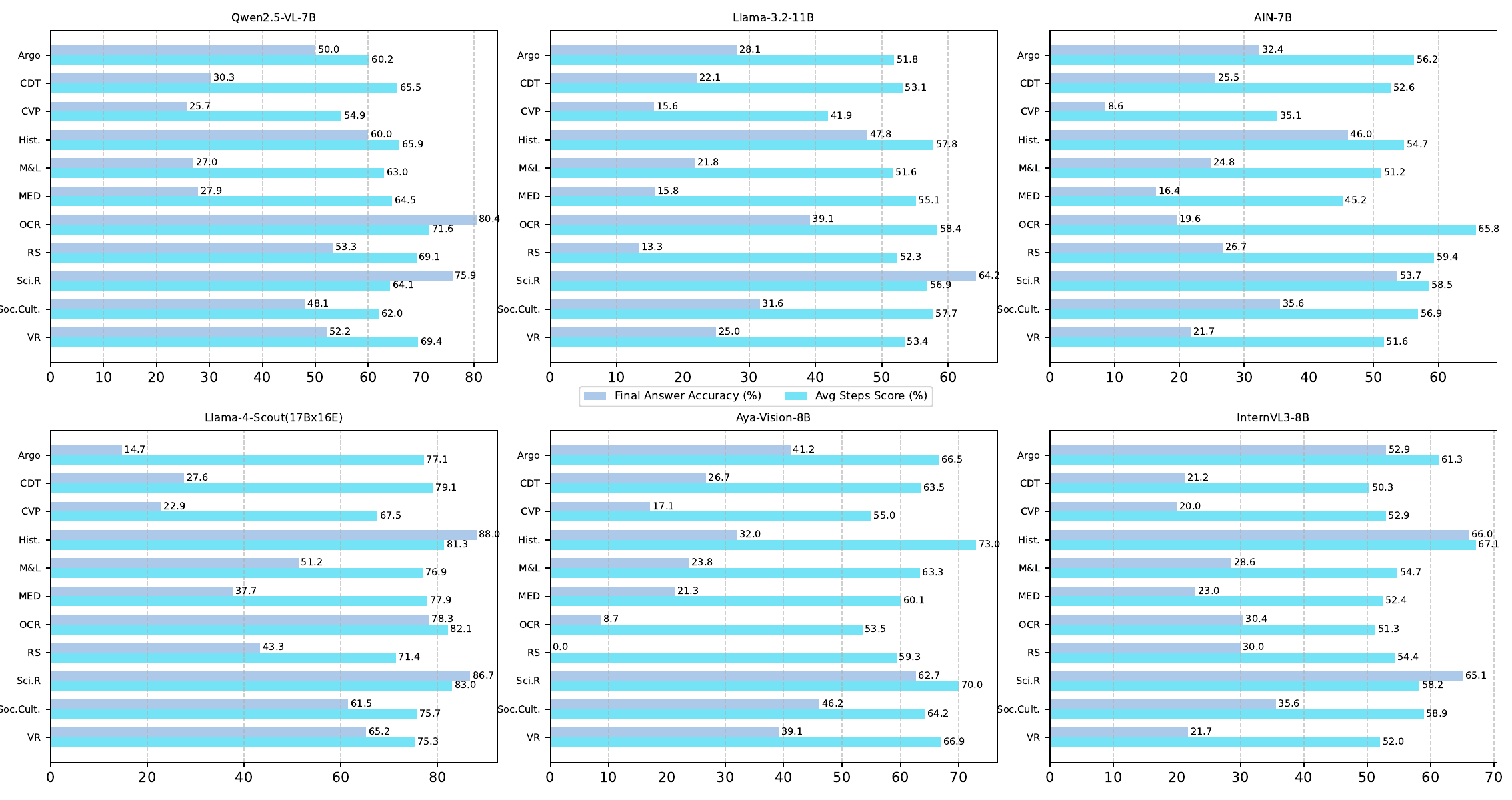}
\vspace{-1.5em}
   \caption{\textbf{Domain-Level Performance of Open-Source Models.}  
Comparison of final answer accuracy and reasoning step scores across ARB domains for six open-source models. LLaMA-4 and AIN perform well in Sci.R and OCR but struggle in RS and VR. Qwen2.5-VL and LLaMA-3.2 show large gaps between reasoning and answers, especially in culturally grounded domains (e.g., Hist., Soc.Cult.). The figure illustrates challenges open models face in Arabic cross-modal reasoning.}
  \label{fig:hcharts_open}
\vspace{-1em}
\end{figure*}

\begin{figure*}[hptb]
\centering  
\includegraphics[width=1.05\textwidth,height=21cm]{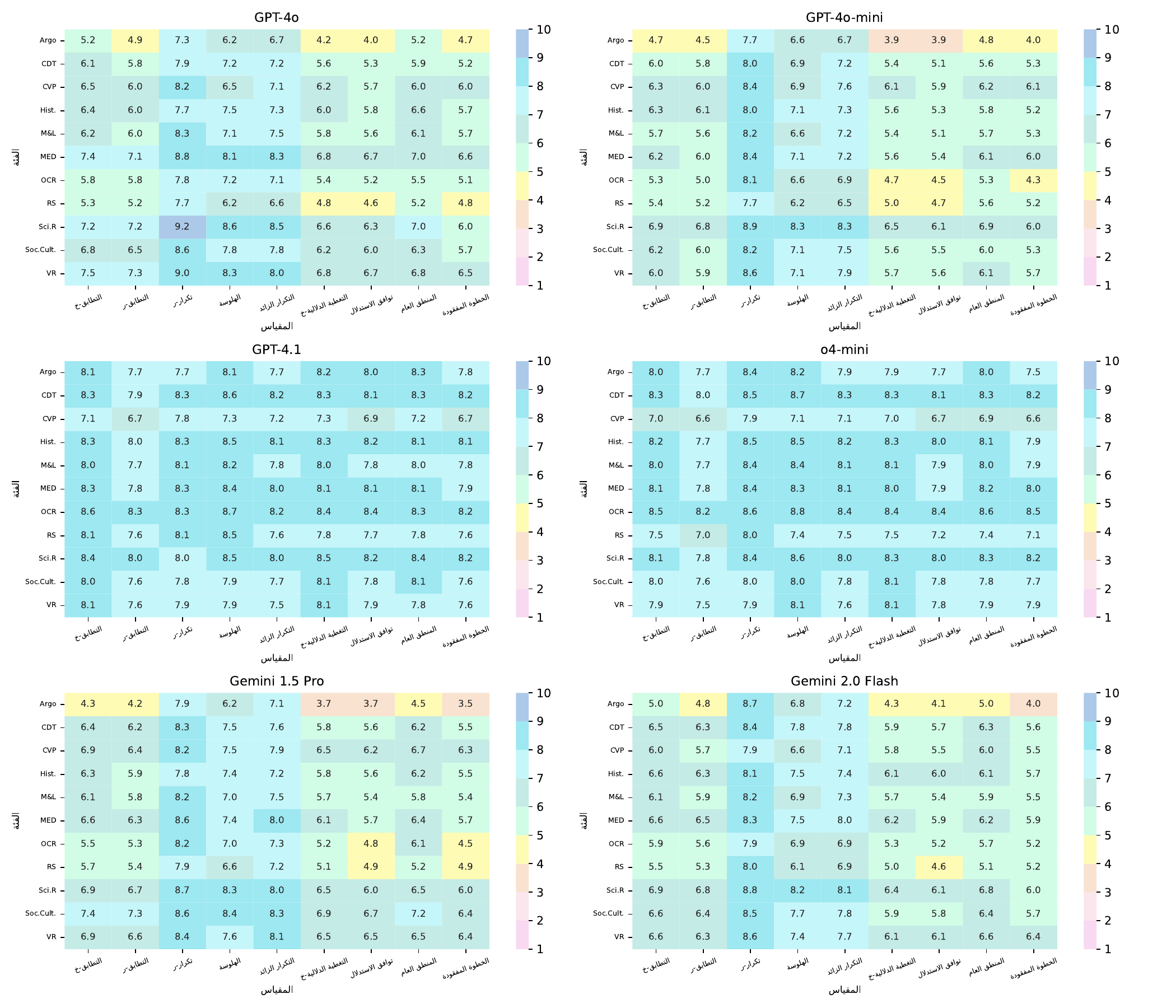}
\vspace{-1.5em}
\caption{\textbf{Stepwise Attribute-Level Evaluation of Closed-Source Models.}  
Heatmaps illustrating the average scores (1–10 scale) across key reasoning attributes—faithfulness, coherence, informativeness, and other diagnostic criteria—within each ARB domain for six closed-source models, based on the LLM-as-Judge framework using Arabic prompts. Models such as GPT-4.1 and o4-mini consistently achieve high scores across most attributes and domains, particularly in Sci.R, CDT, and Hist., indicating strong reasoning reliability. In contrast, performance degrades in perceptual-heavy domains like CVP and RS, where scores drop across multiple attributes. The heatmaps also expose granular inconsistencies—e.g., faithfulness gaps in MED or informativeness variability in Agro—that would be obscured by aggregate metrics. These results emphasize the value of attribute-level evaluation in diagnosing model reasoning quality in Arabic multimodal tasks.}
  \label{fig:heatmap_closed}
  \vspace{-1em}
\end{figure*}

\begin{figure*}[hptb]
\centering  
\includegraphics[width=1.05\textwidth,height=21cm]{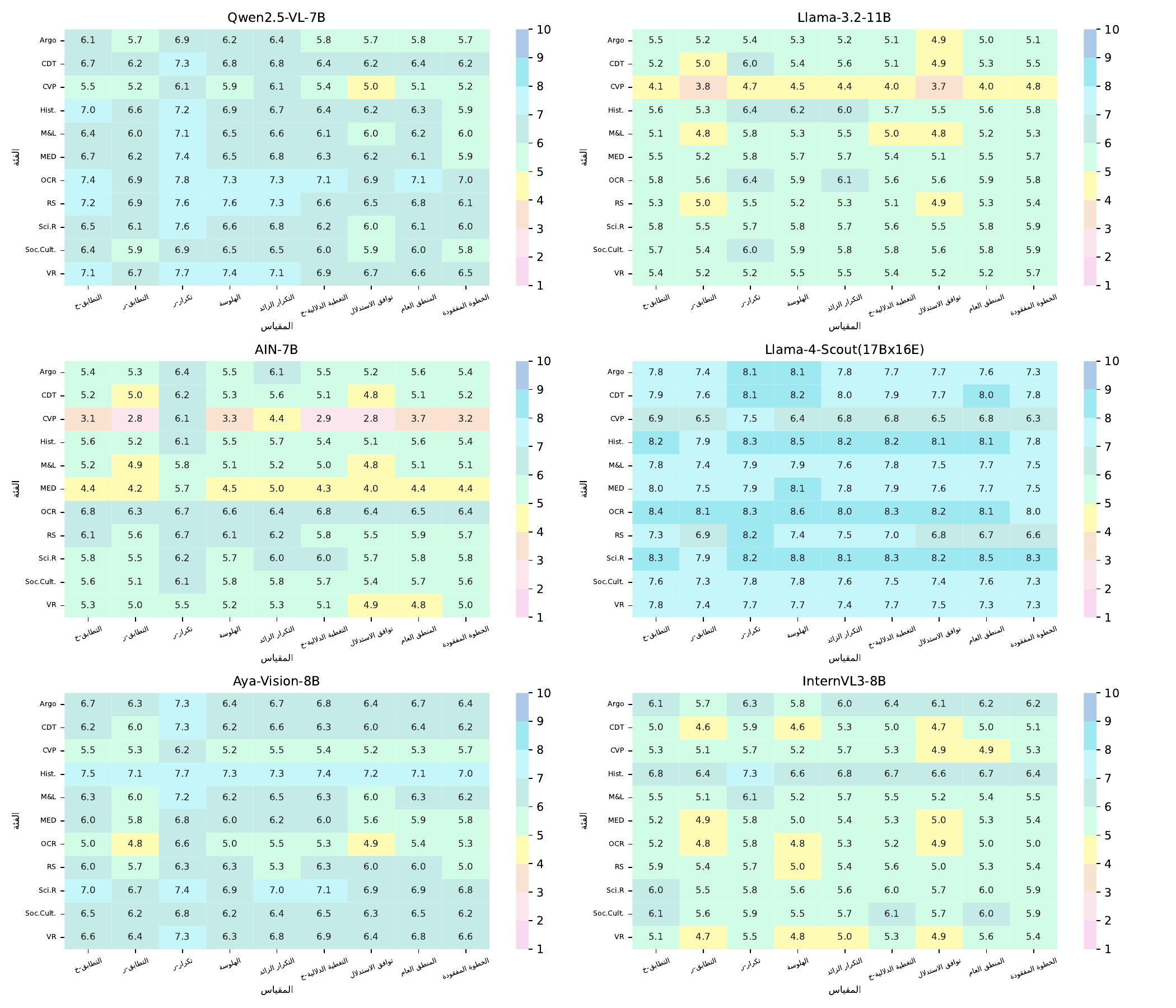}
\vspace{-1.5em}
\caption{\textbf{Stepwise Attribute-Level Evaluation of Open-Source Models.}  
Heatmaps visualizing average attribute-level scores (1–10 scale) across ARB domains for six open-source models, based on the LLM-as-Judge framework using Arabic prompts. Each cell reflects the model's performance across core reasoning dimensions—faithfulness, coherence, informativeness, and error-related factors—per domain. Models such as LLaMA-4 and AIN demonstrate consistent stepwise quality across scientific and OCR tasks, while others like Qwen2.5-VL and LLaMA-3.2 struggle in culturally sensitive or perception-heavy domains (e.g., Hist., Soc.Cult., RS). These results offer fine-grained insight into open-model weaknesses and underscore the importance of domain- and attribute-aware evaluation in Arabic multimodal reasoning tasks.}

  \label{fig:heatmap_open}
  \vspace{-1em}
\end{figure*}

The bar charts (Figures~\ref{fig:hcharts_closed} and~\ref{fig:hcharts_open}) provide an overview of the aggregated scores, while the heat maps (Figures~\ref{fig:heatmap_closed} and~\ref{fig:heatmap_open}) offer a more granular perspective on domain-level performance across individual evaluation metrics. Together, these figures reveal consistent discrepancies between reasoning coherence and final answer correctness, and highlight domain-specific strengths and weaknesses across model types.

\section{Qualitative Examples}
\label{app:qual_sec}

\begin{figure*}[hptb]
\centering  
\includegraphics[width=\textwidth, height=10cm]{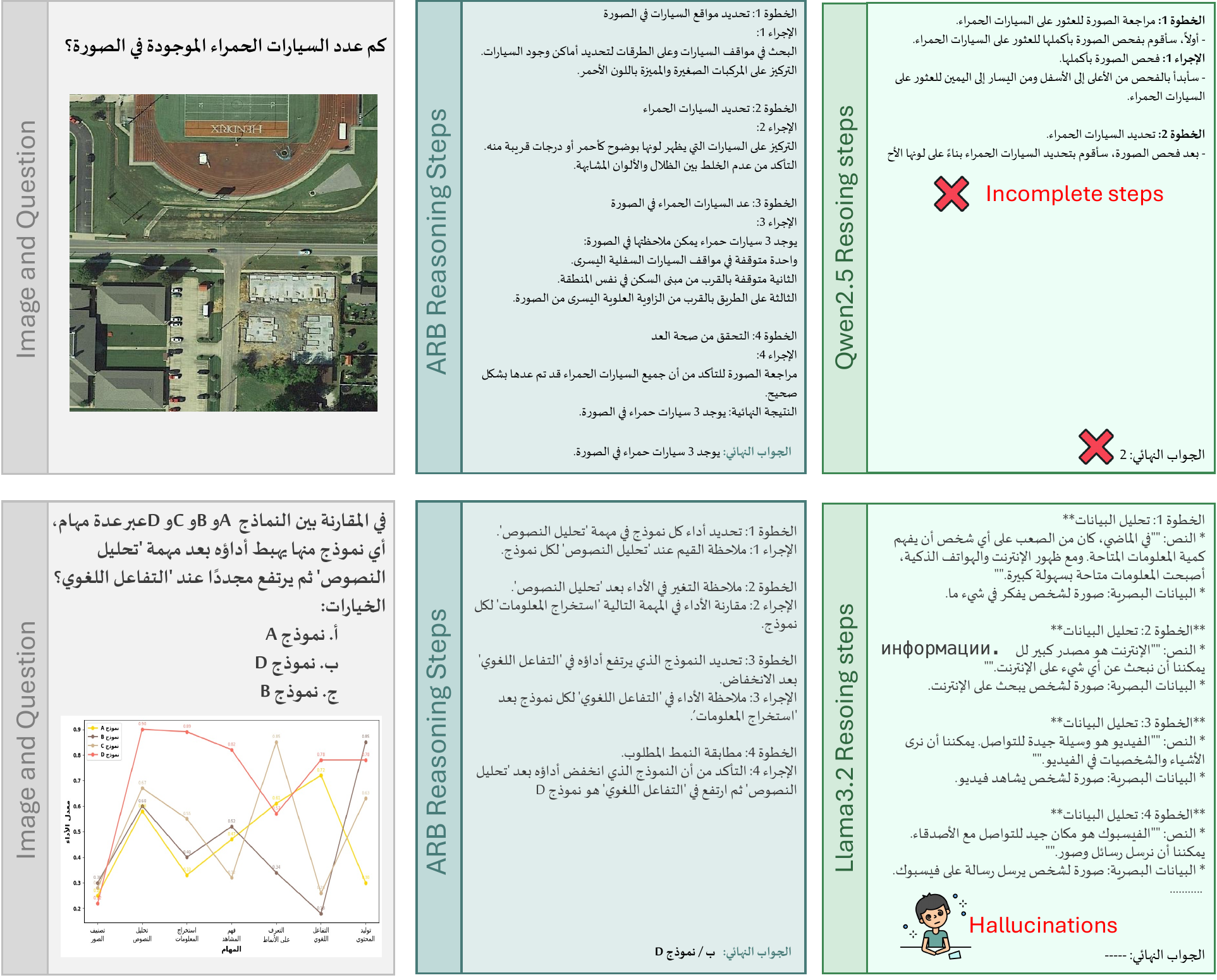}
\vspace{-1.5em}
\caption{\textbf{Qualitative Errors in Open-Source Models.}
This figure showcases common reasoning flaws in open-source LMMs across diverse Arabic multimodal tasks. Errors include incomplete reasoning steps, inconsistent logic, and hallucinated interpretations not grounded in the input. These issues often result in incorrect answers or unreliable outputs, reflecting the challenges open models face in structured Arabic reasoning.}

\label{fig:open_qual}
   \vspace{-0.5em}
\end{figure*}

\begin{figure*}[hptb]
\centering  
\includegraphics[width=\textwidth, height=10cm]{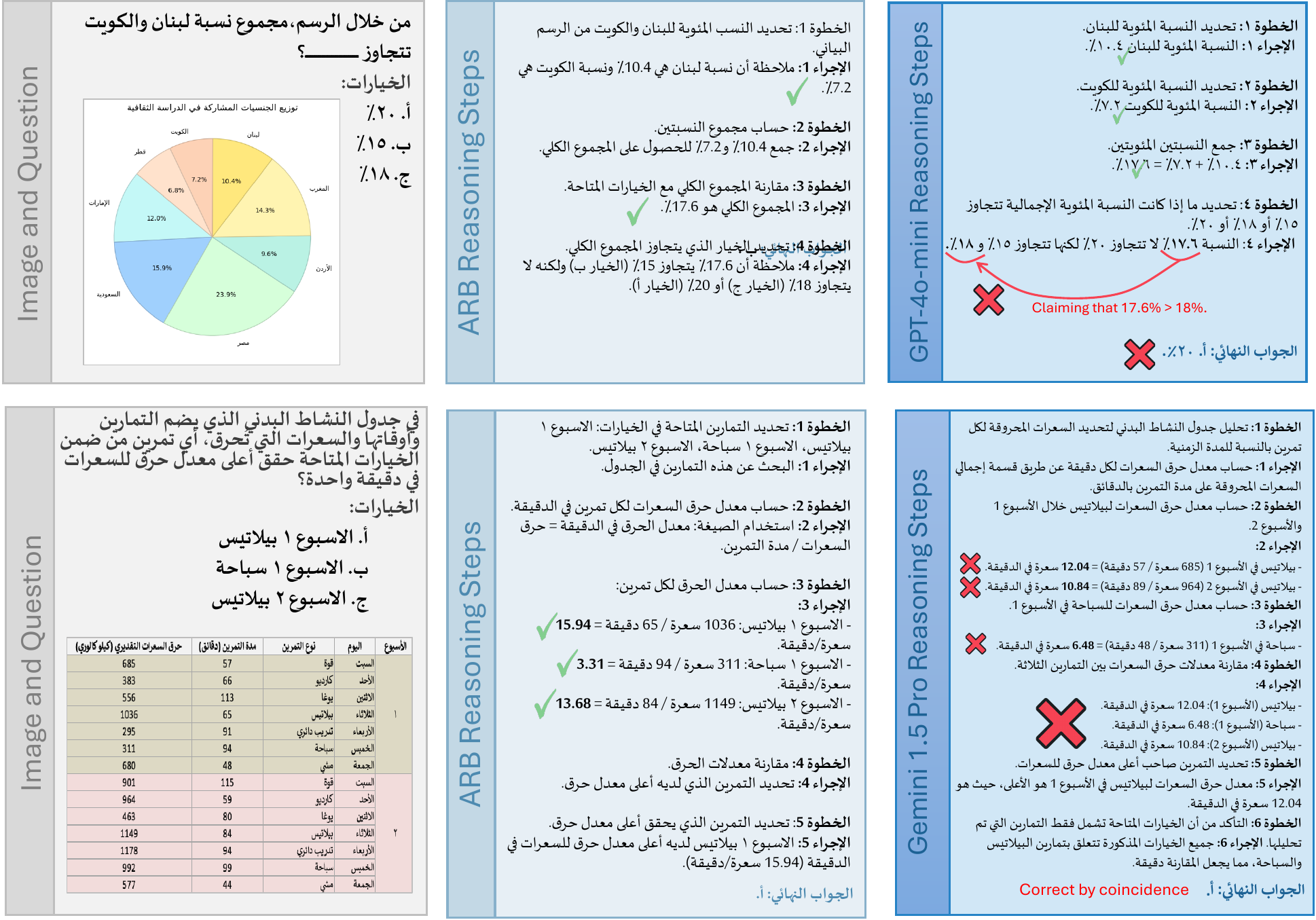}
\vspace{-1.5em}
\caption{\textbf{Qualitative Errors in Closed-Source Models.}
This figure highlights reasoning failures by closed-source LMMs across various Arabic multimodal tasks. Common issues include incorrect numerical comparisons, invalid assumptions, misinterpreted constraints, and logically inconsistent step sequences. These errors often lead to incorrect conclusions despite the appearance of structured reasoning, underscoring the limitations of current closed models when operating in Arabic.}

\label{fig:closed_qual}
   \vspace{-1em}
\end{figure*}

As a further illustration of the quantitative trends discussed in section~\ref{sec:res_analysis}, we present qualitative examples of reasoning failures in both open- and closed-source models (Figures~\ref{fig:open_qual} and~\ref{fig:closed_qual}). These examples reveal persistent issues such as incomplete reasoning chains, hallucinated content, and misapplied constraints across a range of Arabic multimodal tasks. While some outputs appear structurally coherent, they often fail to adhere to task-specific logic or factual correctness. These qualitative insights reinforce the need for Arabic-centric benchmarks like ARB to diagnose and improve model behavior in complex reasoning scenarios.

\section{Data Statistics}
\label{sec:data_stat}

\subsection{Distribution of Reasoning Steps per Sample}
\begin{figure*}[!b]
\centering  
\includegraphics[width=\textwidth, height=7cm]{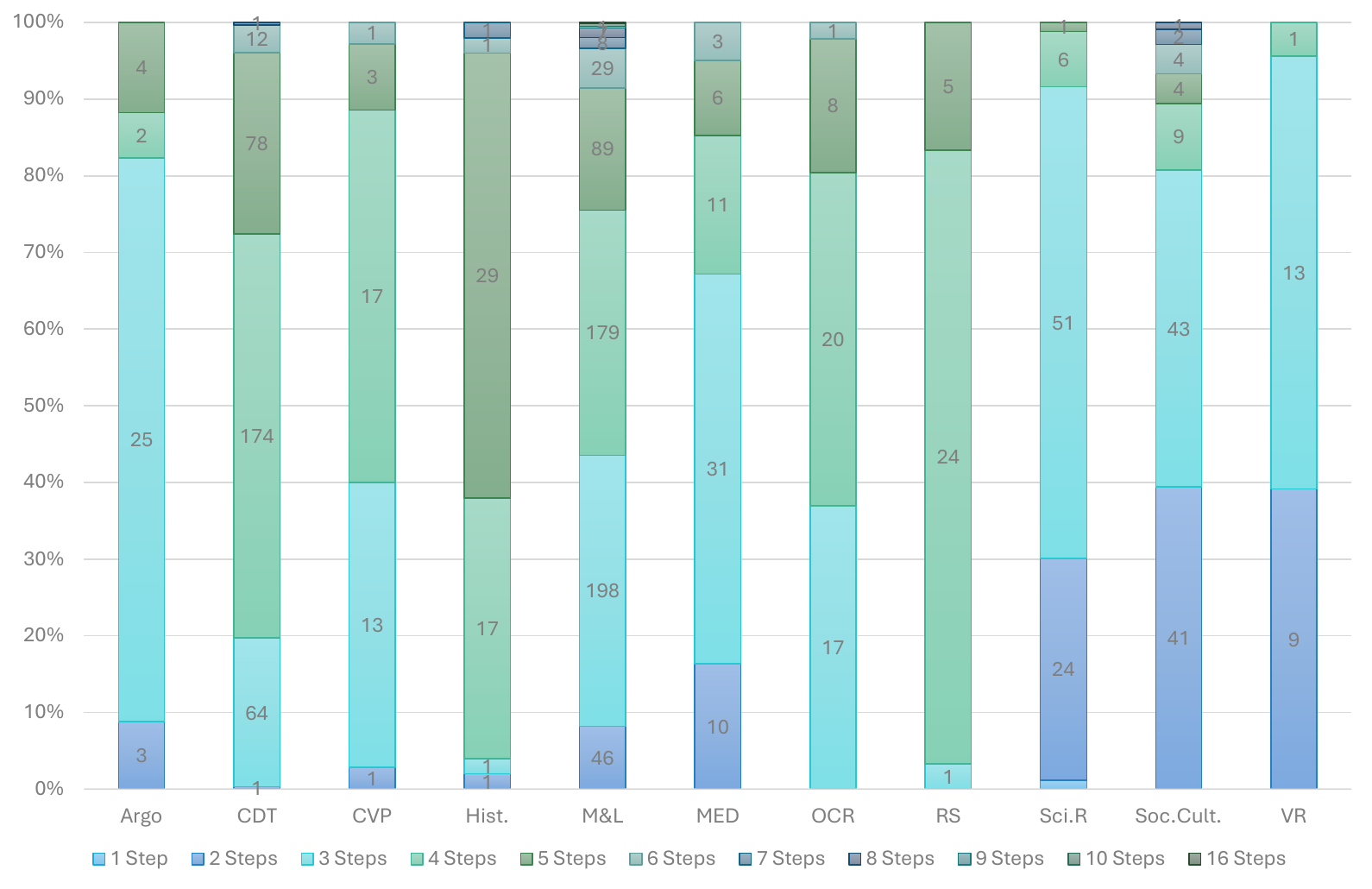}
\vspace{-1.5em}
\caption{\textbf{Step Count Distribution by Domain.} This figure shows the frequency distribution of reasoning steps per sample across the 11 ARB domains. Most domains exhibit a concentration between 2 and 6 steps, with Math \& Logic, History, and Remote Sensing containing a larger share of samples requiring extended reasoning chains.}
\label{fig:cat_step_count}
   \vspace{-1em}
\end{figure*}

To examine the structure of the ARB benchmark across domains, we report key statistical findings. Figure~\ref{fig:cat_step_count} illustrates the distribution of step counts in all ARB entries over their domains, revealing the frequency and variance of the step depth required for the completion of the task.

\subsection{Token Count by Domain}
\begin{figure*}[ht!]
\centering
% Subfigure (a)
    \begin{subfigure}{\textwidth}
        \centering
        \includegraphics[width=\textwidth,height=6cm]{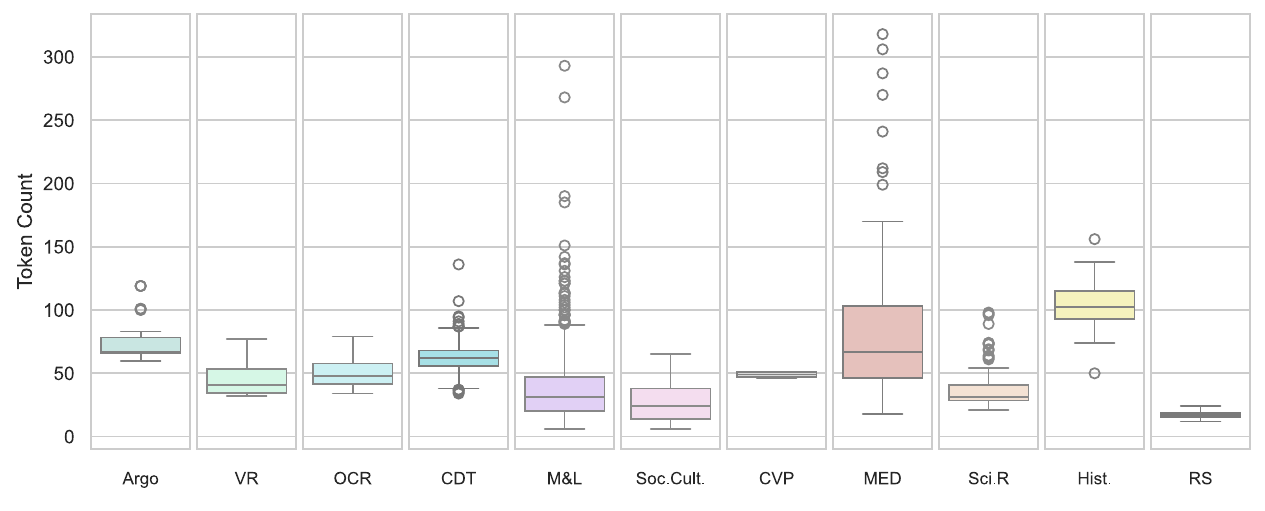}
        \caption{\textbf{Question Token Length Distribution by Domain.} The figure shows the distribution of token counts for questions across different domains in ARB. Domains such as Medical Reasoning (MED) and Historical and Archeological Understanding (Hist.) exhibit higher variability and longer questions, reflecting their inherent complexity.}

        \label{fig:qtoken-a}
    \end{subfigure}
\vspace{0.5em}
% Subfigure (b)
    \begin{subfigure}{\textwidth}
        \centering
        \includegraphics[width=\textwidth,height=6cm]{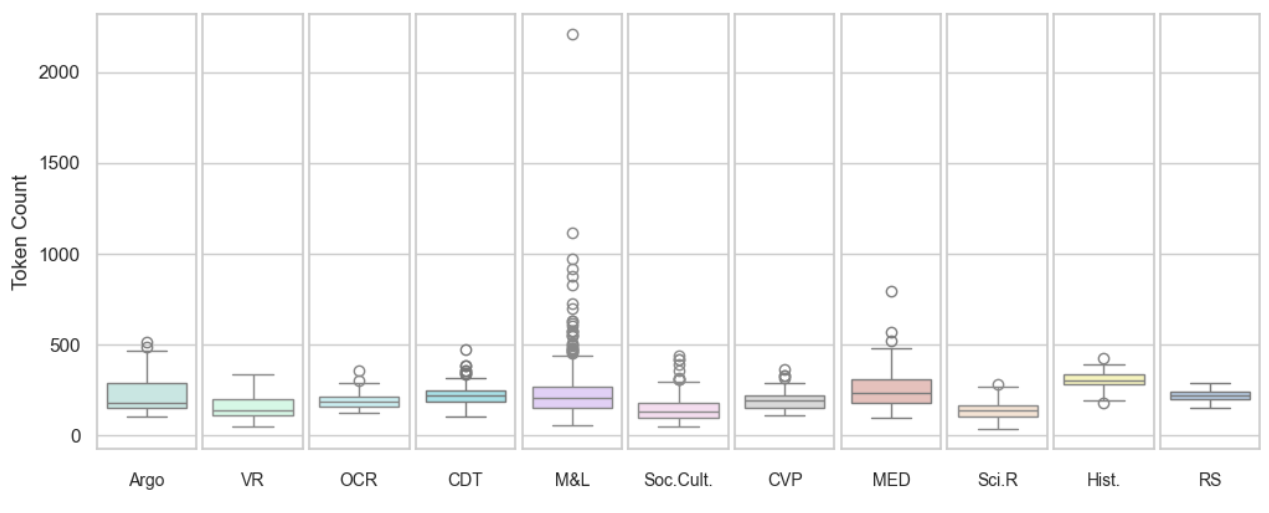} % Change this file
       \caption{\textbf{Reasoning Steps Token Length Distribution by Domain.} The figure presents the distribution of token counts for the generated reasoning steps across domains. Reasoning steps tend to be longer in complex domains such as Medical, Math \& Logic, and Historical \&  Archaeological Understanding (Hist.), highlighting the need for extended multi-hop reasoning.}
        \label{fig:qtoken-b}
    \end{subfigure}
\vspace{-2.5em}
\caption{Question token analysis in ARB: (a) token length by domain, and (b) [describe the second figure].}
\label{fig:Qtoken}
\end{figure*}

Figure~\ref{fig:qtoken-a} shows the distribution of question token lengths across domains. Most questions are relatively concise, but domains such as Medical Reasoning (MED) and Historical and Archaeological Understanding (Hist.) exhibit higher variability and longer lengths. This reflects the inherent complexity and information density required in specialized domains. Similarly, Figure~\ref{fig:qtoken-b} presents the token length distribution of the reasoning steps. These are often longer in domains like Medical Reasoning, Math and Logic (M\&L), and Historical and Archaeological Understanding, indicating the need for more elaborate multi-step reasoning in knowledge-intensive tasks.

\subsection{ Question-to-Reasoning Token Ratio}
\begin{figure*}[t]
\centering  
\includegraphics[width=\textwidth]{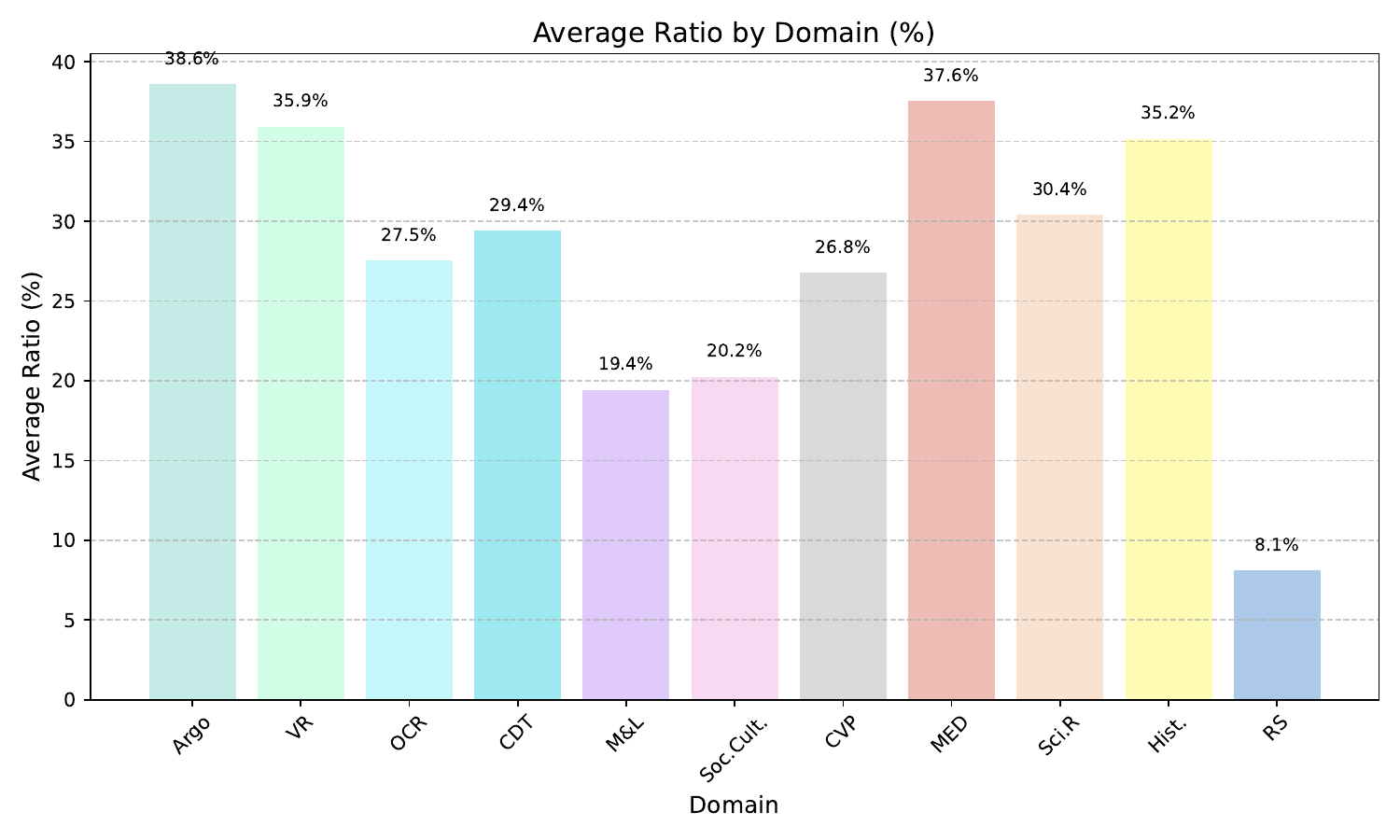}
\vspace{-2em}
  \caption{\textbf{Question-to-Reasoning Token Ratio by Domain.} The figure illustrates the average ratio between question token lengths and reasoning step token lengths across domains. Higher ratios in domains like Argo and MED indicate that these tasks require significantly more elaborate reasoning chains compared to the original question length.}

  \label{fig:ratio}
\end{figure*}

Figure~\ref{fig:ratio} depicts the average ratio of question tokens to reasoning step tokens across domains. Generally, reasoning steps are significantly longer than the original questions, with ratios exceeding 30\% in most cases. Notably, the Medical Reasoning (MED) and Agricultural Image Interpretation (Argo) domains show the highest ratios, suggesting that these tasks demand extensive inferential elaboration beyond the surface-level query.

\subsection{Performance Correlation with Length}
Preliminary analysis indicates that longer reasoning chains are modestly correlated with improved performance in complex domains such as Medical and Scientific Reasoning. However, excessive verbosity does not consistently yield higher accuracy, highlighting the importance of targeted, efficient reasoning over mere length.

\begin{figure}[hptb]
\centering  
\includegraphics[width=0.45\textwidth]{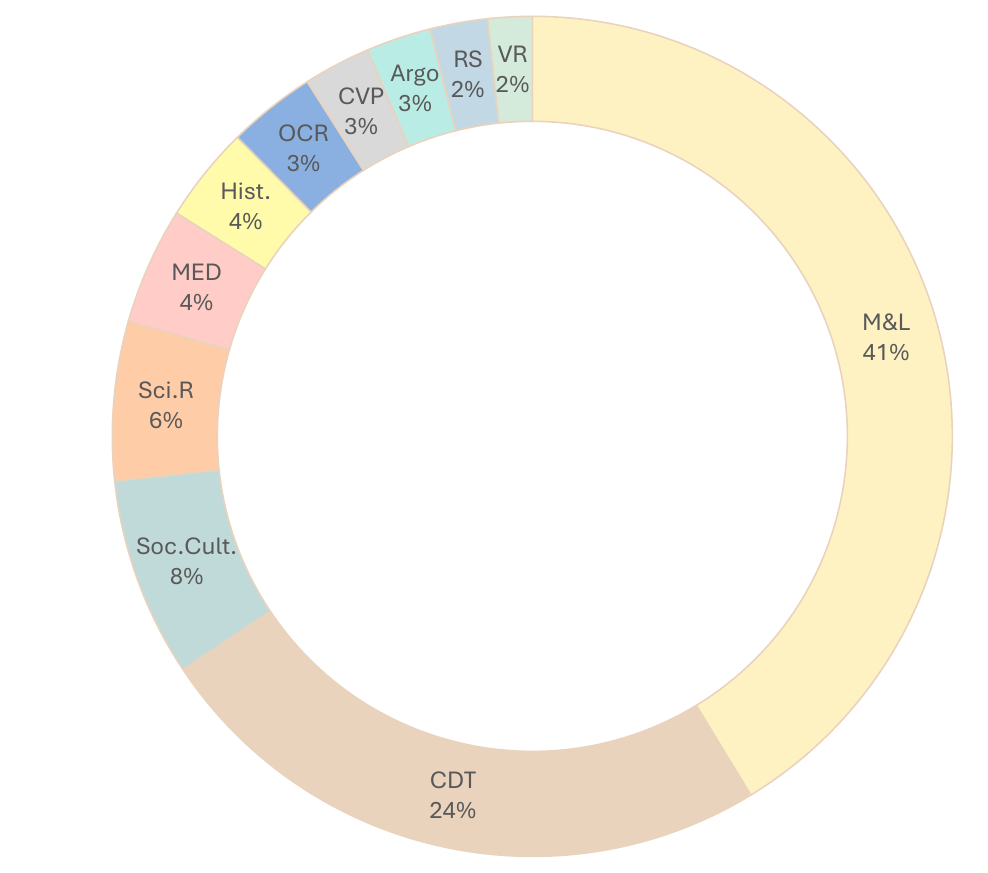}
\vspace{-1em}
\caption{\textbf{Domain Distribution in ARB.} The figure shows the distribution of ARB samples across 11 domains. Math \& Logic (41\%) and Charts, Diagrams, \& Tables (24\%) dominate, reflecting the dataset’s emphasis on structured reasoning. Other domains, including Social \& Cultural, Scientific, and Medical, add thematic diversity.}
\label{fig:domain_dist}
   % \vspace{-1em}
\end{figure}

\subsection{Average Number of Steps and Domain Effects}
On average, domains such as Medical, Scientific Reasoning, and Historical and Archaeological Understanding require a greater number of reasoning steps per question, compared to more straightforward domains like OCR or Remote Sensing (RS). This suggests that scientifically and historically grounded tasks inherently involve deeper multi-hop reasoning, presenting greater challenges for both human annotators and models.

\end{document}